%% file: main.tex
\documentclass{article}

\input{lib/dependencies.tex}
\input{lib/pragmaonce.tex}

\ifdefined\mydraft
\mydraft
\fi

\defaultbibliography{bibl}
\defaultbibliographystyle{apalike-ejor}

\begin{document}

\input{text/frontmatter.tex}

\begin{bibunit}

\input{text/body.tex}

\input{text/acknowledgement.tex}

\input{text/references.tex}

\end{bibunit}

\clearpage
\newpage

\begin{bibunit}

\input{text/supplement.tex}

\input{text/references.tex}
\end{bibunit}

\end{document}

%% file: lib/dependencies.tex
\usepackage{adjustbox}
\usepackage{algorithmic}
\usepackage{amsmath,amssymb,amsfonts}
\usepackage{array}
\usepackage{authblk}
\usepackage{bibunits}
\usepackage{booktabs}
\usepackage{caption}
\usepackage{cite}
\usepackage{colortbl}
\usepackage{csvsimple}
\usepackage{csquotes}
\usepackage{etoolbox}
\usepackage{graphicx}
\usepackage{hyperref,cleveref}
\usepackage{import}
\usepackage{natbib}[sort]
\usepackage{newunicodechar}
\usepackage{rotating}
\usepackage[moderate]{savetrees}
\usepackage{siunitx}
\usepackage{subcaption}
\usepackage{tabularx}
\usepackage{textcomp}
\usepackage{xcolor}
\usepackage{lib/alifeconf}
\usepackage{orcidlink}
\usepackage{pdflscape,longtable}
\usepackage{listings}
\usepackage{placeins}
\usepackage{pythonhighlight}

\newunicodechar{🧬}{\includegraphics[height=\fontcharht\font`A]{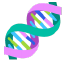}}
\newunicodechar{🦠}{\includegraphics[height=\fontcharht\font`A]{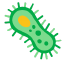}}

%% file: lib/pragmaonce.tex
\makeatletter
\let\pragma@iinput=\@iinput
\def\@iinput#1{\xdef\@pragmafile{#1}\pragma@iinput{#1} }
\def\@pragmafile{default}
\def\pragmaonce{%
   \csname pragma@\@pragmafile\endcsname
   \global\expandafter\let \csname pragma@\@pragmafile\endcsname =  
}
\makeatother

%% file: text/frontmatter.tex
\title{ Trackable Agent-based Evolution Models at Wafer Scale}

\author[1,2,3,*]{Matthew Andres Moreno\orcidlink{0000-0003-4726-4479}}
\author[4]{Connor Yang\orcidlink{0009-0004-1240-2362}}
\author[5,6]{Emily Dolson\orcidlink{0000-0001-8616-4898}}
\author[1,2]{Luis Zaman\orcidlink{0000-0001-6838-7385}}

\affil[.  ]{\vspace{-4.4ex}}
\affil[1]{Department of Ecology and Evolutionary Biology, University of Michigan, Ann Arbor, United States}
\affil[2]{Center for the Study of Complex Systems, University of Michigan, Ann Arbor, United States}
\affil[3]{Michigan Institute for Data Science, University of Michigan, Ann Arbor, United States}
\affil[4]{Undergraduate Research Opportunities Program, University of Michigan, Ann Arbor, United States}
\affil[5]{Department of Computer Science and Engineering, Michigan State University, East Lansing, United States}
\affil[6]{Program in Ecology, Evolution, and Behavior, Michigan State University, East Lansing, United States}
\affil[*]{corresponding author: \textit{morenoma@umich.edu}}

\maketitle

%% file: text/body.tex
\input{text/body/abstract.tex}

\input{text/body/introduction.tex}

\input{text/body/methods.tex}

\input{text/body/results-and-discussion.tex}

\input{text/body/conclusion.tex}

%% file: text/body/abstract.tex
\begin{abstract}
\vspace{-2ex}
Continuing improvements in computing hardware are poised to transform capabilities for \textit{in silico} modeling of cross-scale phenomena underlying major open questions in evolutionary biology and artificial life, such as transitions in individuality, eco-evolutionary dynamics, and rare evolutionary events.
Emerging ML/AI-oriented hardware accelerators, like the 850,000 processor Cerebras Wafer Scale Engine (WSE), hold particular promise.
However, many practical challenges remain in conducting informative evolution experiments that efficiently utilize these platforms' large processor counts.
Here, we focus on the problem of extracting phylogenetic information from agent-based evolution on the WSE platform.
This goal drove significant refinements to decentralized \textit{in silico} phylogenetic tracking, reported here.
These improvements yield order-of-magnitude performance improvements.
We also present an asynchronous island-based genetic algorithm (GA) framework for WSE hardware.
Emulated and on-hardware GA benchmarks with a simple tracking-enabled agent model clock upwards of 1 million generations a minute for population sizes reaching 16 million agents.
This pace enables quadrillions of agent replication events a day.
We validate phylogenetic reconstructions from these trials and demonstrate their suitability for inference of underlying evolutionary conditions.
In particular, we demonstrate extraction, from wafer-scale simulation, of clear phylometric signals that differentiate runs with adaptive dynamics enabled versus disabled.
Together, these benchmark and validation trials reflect strong potential for highly scalable agent-based evolution simulation that is both efficient and observable.
Developed capabilities will bring entirely new classes of previously intractable research questions within reach, benefiting further explorations within the evolutionary biology and artificial life communities across a variety of emerging high-performance computing platforms.
\end{abstract}

%% file: text/body/introduction.tex
\section{Introduction}

A quintessential characteristic of computational artificial life experiments is the near total malleability of the simulacrum \citep{pattee1989simulations}.
Indeed, exploration of arbitrary possibilities `as they could be' is the core of artificial life's role as a tool for inquiry \citep{langton1997artificial}.
Such near-limitless freedom to realize arbitrary system configurations, however, can obscure an intrinsic limitation of most computational artificial life work: scale.

Take, for instance, the Avida platform, which instantiates populations of self-replicating computer programs for evolution experiments.
When running on a single CPU, this system can support about 20,000 generations per day, or about two hundred million individual replication cycles daily \citep{ofria2009artificial}.
By way of comparison, \textit{E. coli} populations within individual flasks of the Lenski Long-Term Evolution Experiment undergo only six doublings per day, meaning their generations take orders of magnitude longer than Avidians \citep{good2017dynamics}.
(In continuous culture, though, the rate can be c. 72 generations per day.)
Indeed, such capability for fast generational turnover has been a key motivation for using artificial life systems to study evolution.
However, the effective population size of flasks in the Long-Term Evolution Experiment is orders of magnitude larger than Avida's population size: 30 million vs. 10,000.
Consequently, these systems actually exhibit a similar number of replication events per day.
This pattern of dramatically faster generation times than those observed in nature and dramatically smaller populations largely generalizes across artificial life systems.
Of course, any such comparisons should also note profound discrepancies between the genetic, phenotypic, and environmental richness of biological organisms and ALife models.

Crucially, however, the scale of population size can greatly impact subjects of artificial life research, like transitions in individuality, ecological dynamics, and rare evolutionary innovations \citep{taylor2016open,dolson2021digital,taylor2019evolutionary}.
Cross-scale dynamics are also crucial to many key real-world challenges.
For example, in evolutionary epidemiology, interactions between within-host infection dynamics and population-level epidemiological patterns determine the evolutionary trajectory of the population \citep{schreiber2021cross}.
However, because capabilities of current silicon-based processors are not expected to improve markedly in the foreseeable future \citep{sutter2005free}, the scale-up necessary to progress on these key frontiers will demand many-processor computation.
Application of parallel and distributed computation, however, imposes compromises to the convenience, flexibility, observability, interpretability, total reliability, and perfect replicability enjoyed under classical centralized, serial models of computation.
Encouragingly, these challenges are already implicit to much of biology; the productivity of research involving natural organisms evidences that they are surmountable and even hints at strategies that can be used to solve them.
Here, we explore alignment of digital evolution to HPC accelerator hardware at the extreme cutting edge of massively distributed computation, and use techniques inspired by those applied to natural organisms to mitigate limitations of distributed computation with respect to tracking phylogenies.

\subsection{Progress Toward Scale-up in Artificial Life}

Achieving highly scalable artificial life and digital evolution systems involves two distinct engineering considerations.
First, as with any high-performance scientific computing, careful design is required to appropriately divvy computation and communication workloads across available hardware.
Second, given the exceptionally discretionary nature of artificial life modeling, we can intentionally tailor simulation semantics to suit underlying hardware capabilities.
Ackley's ongoing work with the T2 Tile Project and ULAM exemplifies a strong synthesis of this engineering duality \citep{ackley2016ulam}.
At the level of simulation semantics, Ackley formulates update procedures in terms of local interactions between discrete, spatially situated particles.
This design provides for efficient one-to-one mapping between simulation space and hardware components, minimizing requirements for intra-hardware connectivity and preventing global impacts from on-the-fly augmentations or reductions of available hardware.
The ULAM framework then ties into implementation-level infrastructure necessary to accomplish performant, best-effort lock/release of spatial event windows spanning bordering hardware units \citep{ackley2013movable}.
Ackley's work is distinguished, in fact, in dipping to a yet-lower level of abstraction and tackling design of bespoke, modular distributed processing hardware \citep{ackley2011homeostatic,ackley2023robust,livingcomputationSFBSanta}.

Several additional digital evolution projects have made notable headway in synthesizing artificial life models with sophisticated, scalable technical backing, achieving rich interactions among numerous parallelized simulation components.
Harding demonstrated large-scale cellular automata-based artificial development systems, achieved through GPU-parallelized instantiations of a genetic program  \citep{harding2007fast_ieee}.
Early work by Ray with Network Tierra used an island model to distribute digital organisms across federated servers, with migration handled according to the real-time latencies and topology of the underlying network \citep{ray1995proposal}.
More recently, Heinemann's continuation of the ALIEN project has leveraged GPU acceleration to achieve spectacularly elaborate simulations with rich interactions between numerous spatially-situated soft body agents \citep{heinemann2008artificial}.
Likewise, the Distributed Hierarchical Transitions in Individuality (DISHTINY) project has incorporated event-driven agent-agent interaction schemes amenable to best-effort, asynchronous interlocution \citep{moreno2022exploring,moreno2021conduit}.
GPU-first agent-based modeling (ABM) packages like Flame GPU also tackle this problem of hardware-simulacrum matching, albeit framed at a higher level of abstraction \citep{richmond2010high}.
Beyond ALife, broader realms of application-oriented evolutionary computation have folded in with many-processor computation, most commonly through island-model and director-worker evaluation paradigms \citep{abdelhafez2019performance,cantu2001master}.

\subsection{Untapped Emerging Hardware}

In retrospect, connectionist artificial intelligence turns out to have been profoundly scale-dependent.
The utility and ubiquity of ANNs have exploded in tandem with torrential growth in training set sizes, parameter counts, and training FLOPs \citep{marcus2018deep}.
Recruitment of multi-GPU training for image classification, requisite particular accommodating adjustments to the underlying deep learning architecture, is commonly identified as the watershed moment to this transformation
 \citep{krizhevsky2012imagenet}.
Commercial investment in AI capabilities then set in motion a virtuous cycle of further-enabling hardware advances \citep{jouppi2017datacenter}.
Indeed, the scaling relationship between deep learning and training resources has itself become a major area of active study, with expectation for this virtuous cycle to continue through the foreseeable future \citep{kaplan2020scaling}.

A major upshot of the deep learning race is the emergence of spectacularly capable next-generation compute accelerators \citep{zhang2016cambricon,emani2021accelerating,jia2019dissecting,medina2020habana}.
Although tailored expressly to deep learning workloads, these hardware platforms represent an exceptional opportunity to leapfrog progress on grand challenges in artificial life.
The emerging class of fabric-based accelerators, led by the 850,000 core Cerebras CS-2 Wafer-Scale Engine (WSE) \citep{lauterbach2021path,lie2022cerebras}, holds particular promise as a vehicle for artificial life models.
This architecture interfaces multitudinous processing elements (PEs) in a physical lattice, with PEs executing independently with private on-chip memory and interacting locally through a network-like interface.

In this work, we explore how such hardware might be recruited for large-scale digital evolution, demonstrating a genetic algorithm implementation tailored to the dataflow-oriented computing model of the CS-2 platform.
Indeed, rapid advances in the capability of accelerator devices, driven in particular by market demand for deep learning operations, are anticipated to drive advances in agent-based model capabilities \citep{perumalla2022computer}.
The upcoming CS-3 chip, for instance, supports clustering potentially thousands of constituent accelerators \citep{moore2024cerebras}.

\subsection{Maintaining Observability}

Orthogonalities between the fundamental structure and objectives of AI and artificial life methods will complicate any effort to requisition AI hardware for artificial life purposes.
In common use, deep learning operates as a black box medium \citep{loyola2019black} (but not always \citep{mahendran2015understanding}).
This paradigm de-emphasizes accessibility of inner state.
In contrast, artificial life more often functions as a tool for inquiry.
This goal emphasizes capability to observe and interpret underlying simulation state \citep{moreno2023toward,horgan1995complexity}.
(A similar argument holds for ALife work driven by artistic objectives, as well.)

Unfortunately, scale complicates simulation observability.
It is not uncommon for the volume and velocity of data streams from contemporary simulation to outstrip hardware bandwidth and storage capacity \citep{osti_1770192}.
Extensive engineering effort will be required to ensure large-scale simulation retains utility in pursuing rigorous hypothesis-driven objectives.

Here, we confront just a single aspect of simulation observability within distributed evolutionary simulation: phylogenetic history (i.e., evolutionary ancestry relationships).
Phylogenetic history plays a critical role in many evolution studies, across study domains and \textit{in vivo} and \textit{in silico} model systems alike \citep{faithConservationEvaluationPhylogenetic1992,STAMATAKIS2005phylogenetics,frenchHostPhylogenyShapes2023,kim2006discovery,lewinsohnStatedependentEvolutionaryModels2023a,lenski2003evolutionary,moreno2021case}.
Phylogenetic analysis can trace the history of notable evolutionary events (e.g., extinctions, evolutionary innovations), but also characterize more general questions about the underlying mode and tempo of evolution \citep{moreno2023toward,hernandez2022can,shahbandegan2022untangling,lewinsohnStatedependentEvolutionaryModels2023a}.
Particularly notable, recent work has used comparison of observed phylogenies against those produced under differing simulation conditions to test hypotheses describing underlying dynamics within real-life evo-epidemiological systems \citep{giardina2017inference,voznica2022deep}.
Additionally, \textit{in silico}, phylogenetic information can even serve as a mechanism to guide evolution in application-oriented domains \citep{lalejini2024phylogeny,lalejini2024runtime,murphy2008simple,burke2003increased}.

\subsection{Decentralized Phylogenetic Tracking}

\input{fig/schematic}

Most existing artificial life work uses centralized tracking to maintain an exact, complete record of phylogenetic history comprising all parent-offspring relationships that have existed over the course of a simulation \citep{ray1992evolution,bohm2017mabe,de2012deap,garwood2019revosim,godin2019apoget,dolson2024phylotrackpy}.
Typically, records of extinct lineages are pruned to prevent memory bloat \citep{moreno2024analysis}.
Although direct tracking is well suited to serial simulation or centralized controller-worker schemes, runtime communication overheads and sensitivity to data loss impede scaling to highly distributed systems --- particularly those with lean memory capacity like the Cerebras WSE \citep{moreno2024analysis}.
To overcome this limitation, we have developed reconstruction-based approaches to \textit{in silico} phylogenetic tracking \citep{moreno2022hereditary}.
These approaches require no centralized data collection during simulation runtime; instead, they use \textit{post hoc} comparisons among end-state agent genomes to deduce approximate phylogenetic history --- akin to how DNA-based analyses describe natural history.
Figure \ref{fig:runtime-posthoc-schematic} summarizes this reconstruction-based strategy.

Although analogous work with natural biosequences is notoriously challenging and data-intensive \citep{neyman1971molecular,lemmon2013high},
the recently-developed hereditary stratigraphy annotation architecture is explicitly designed for fast, accurate, and data-lean reconstruction.
Designed to attach on underlying replicators as a neutral annotation (akin to noncoding DNA), it is a general-purpose technique potentially applicable across diverse study domains \citep{liben2008tracing,cohen1987computer,friggeri2014rumor}.
In a stroke of convergent thinking, \citet{ackley2023robust} reports use of ``bar code'' annotations on his self-replicators to achieve a measure of coarse-grained lineage tracing.

\subsection{Contributions}

In this paper, we report new software and algorithms that harness the Cerebras Wafer-Scale Engine to enable radically scaled-up agent-based evolution while retaining key aspects of observability necessary to support hypothesis-driven computational experiments.
Implementation comprises two primary aspects:
\begin{enumerate}
  \item an asynchronous island-based genetic algorithm (GA) suited to the memory-scarce, highly-distributed, data-oriented WSE architecture, and
  \item a fundamental reformulation of hereditary stratigraphy's core storage and update procedures to achieve fast, simple, resource-lean annotations compatible with unconventional, resource-constrained accelerator and embedded hardware like the WSE.
\end{enumerate}

Both are implemented in Cerebras Software Language (CSL) and validated using Cerebras' SDK hardware emulator.
We use benchmark experiments to evaluate the runtime performance characteristics of the new hereditary stratigraphy algorithms in isolation and in the integrated context providing tracking-enabled support for the island-model GA.
In conjunction, we report emulated and on-device trials that validate phylogenetic reconstructions and demonstrate their suitability for inference of underlying evolutionary conditions.

Results from both experiments are promising.
We find that new surface-based algorithms greatly improve runtime performance.
Scaled-down emulator benchmarks and early on-hardware trials indicate potential for simple agent models --- with phylogenetic tracking enabled --- to achieve on the order of quadrillions of agent replication events a day at full wafer scale, with support for population sizes potentially reaching hundreds of millions.
Further, using proposed techniques, phylogenetic analyses of simulations spanning hundreds of thousands of PEs succeed in detecting differences in adaptive dynamics between alternate simulation conditions.

%% file: fig/schematic.tex
\begin{figure}
  \vspace{2ex}
    \centering
  \includegraphics[width=0.8\linewidth]{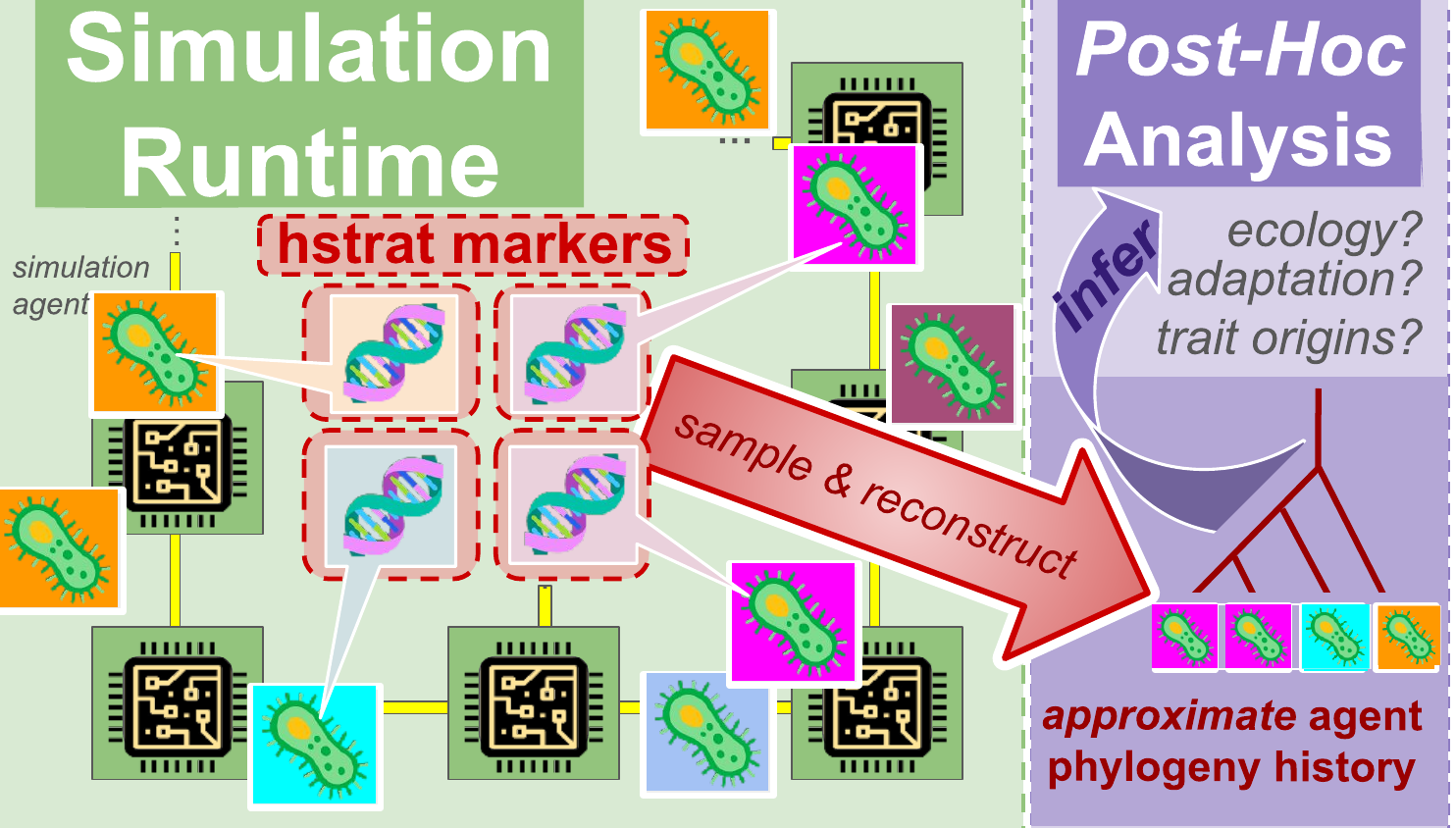}
  \vspace{-1.5ex}
    \caption{\textbf{Strategy for trackable distributed evolution simulation.}
    \footnotesize %
    Hstrat markers (🧬) attached to replicating agents (🦠) in a many-CPU runtime (left panel) enable \textit{post hoc} estimation of relatedness between lineages, enabling approximate phylogenetic reconstruction (right panel).
    }
    \label{fig:runtime-posthoc-schematic}
\vspace{-0.2in}
\end{figure}

%% file: text/body/methods.tex
\section{Methods} \label{sec:methods}

\input{fig/surf-algorithms}

This section recounts the core mechanics of hereditary stratigraphy methods for phylogenetic tracking and describes new lightweight ``surface'' data structures used in this work.
Then, we describe the asynchronous compute-communicate implementation strategy behind the testbed island-based genetic algorithm used for WSE validation and benchmarking experiments.

\subsection{Distributed Phylogenetic Tracking}

Natural history of biological life operates under no extrinsic provision for interpretable record-keeping, yet efforts to study it have proved immensely fruitful.
This fact bodes well that scaled-up simulation studies can succeed as a platform for rich hypothesis-driven experiments despite potential sacrifices to aspects of observability currently enjoyed with centralized, storage-rich processing.
Indeed, observational and analytical strategies already developed to confront limitations in biological data can solve, or at least guide, work with massively distributed evolution simulations.

In biology, mutational drift encodes ancestry information in DNA genomes.
Our proposed methods are analogous, with ancestry information captured within genomes themselves rather than external tracking.
Phylogenetic history can then be estimated after the fact, as Figure \ref{fig:runtime-posthoc-schematic} depicts.
This strategy reduces runtime communication and is resilient to germane modes of data loss (e.g., dropped messages, hardware crash-out).
Note that, in addition to sampling end-state genomes, extracting ``fossil'' genomes over the course of simulation can allow lineages that end up going extinct to be included in phylogenetic reconstruction.

Recent work introducing \textit{hereditary stratigraphy} (hstrat) methodology has explored how best to organize genetic material to maximize reconstruction quality and minimize memory footprint \citep{moreno2022hstrat, moreno2022hereditary}.
Hstrat material can be bundled with agent genomes in a manner akin to non-coding DNA, entirely neutral with respect to agent traits and fitness.

The hereditary stratigraphy algorithm associates each generation along individual lineages with an identifying ``fingerprint'' value, referred to as a differentia.
On birth, offspring receive a new differentia value and append it to an inherited chronological record of past values, each corresponding to a generation along that lineage.
Under this scheme, mismatching differentiae can be used to delimit the extent of common ancestry.
This semantic streamlines \textit{post hoc} phylogenetic reconstruction to a simple trie-building procedure
\citep{moreno2024analysis}.

To save space, differentia may be pruned away.
However, care must be taken to ensure retention of checkpoint generations that maximize coverage across evolutionary history.
Reducing the number of bits per differentia can also provide many-fold memory space savings.
These savings can be re-invested to increase the quantity of differentia retained, improving the density of record coverage across elapsed generations.
The cost of this shift is an increased probability of spurious differentia value collisions, which can make two lineages appear more closely related than they actually are.
We anticipate that most use cases will call for differentia sized on the order of a single bit or a byte.
Indeed, single-bit differentiae have been shown to yield good quality phylogenies using only 96 bits per genome \citep{moreno2024guide}.

Small differentia size intensifies need for a lean data structure to back differentia record management.
Shrinking differentia to a single bit would be absurd if each is accompanied by a 32-bit generational timestamp.
To delete timestamps, though, we need means to recalculate them on demand.
As such, all described algorithms can deduce timestamps of retained differentia solely from their storage index and the count of elapsed generations.

Lastly, design of hstrat annotations must also consider how available storage space should be allocated across the span of history.
In one possible strategy, which we call \textit{\textbf{``steady'' policy}}, retained time points would be distributed evenly across history.
By contrast, under \textit{\textbf{``tilted'' policy}} more recent time points are preferred \citep{han2005stream,zhao2005generalized}.
Note that prior hereditary stratigraphy work refers to them instead as ``depth-proportional'' and ``recency-proportional resolution'' \citep{moreno2022hereditary}.
Comparisons of reconstruction quality have shown that tilted policy gives higher quality reconstructions from the same amount of reconstruction space in most --- but not all --- circumstances \citep{moreno2024guide}.
This pattern follows an intuition that high absolute precision is more useful to resolving recent events than more ancient ones.
In practice, it may be desirable to use a hybrid approach that allocates half of available annotation space to each strategy \citep{moreno2024guide}.
The bottom panels of Figure \ref{fig:surf-algorithms} contrast steady versus tilted behaviors.

\subsection{Surface-based Hereditary Stratigraphy Algorithms}

\input{fig/surf-vs-column-schematic}

At the outset of this project, several problematic aspects of porting existing hereditary stratigraphy algorithms to the WSE became apparent.
Issues stemmed, in part, from a fundamental feature of these algorithms: organization of retained strata in contiguous, sorted order.
This design imposes various drawbacks:
\begin{itemize}
\item \textit{wasted space}: an annotation size cap can be guaranteed, but a percentage of available space typically goes unused;
\item \textit{high-level feature dependencies:} in places, existing column code uses complex data structures with dynamically allocated memory to perform operations like set subtraction; and
\item \textit{annotation size scaling:} maintaining sorted order among differentia can take time linearly proportional to annotation size.
\end{itemize}

\noindent
Addressing these concerns required fundamental reformulation of hereditary stratigraphy conventions.
To this end, we introduce a constant-time indexing scheme that maps each lineage's fingerprint stream directly onto a fixed-width memory buffer.
Differentia pruning occurs implicitly, as a result of resident differentia being overwritten by new placements.
As before, care is taken to guarantee temporally representative collections of resident differentiae within the memory buffer over elapsed time.
In this new \textit{\textbf{``surface''-based approach}}, differentiae remain in the same buffer position until they are overwritten --- as opposed to the existing \textit{\textbf{``column''-based approach}}, where they are kept in sorted order.
Figure \ref{fig:surf-vs-column-schematic} contrasts the two approaches.

In the course of this work, we developed both steady- and tilted-policy surface-based hereditary stratigraphy algorithms.
Figure \ref{fig:surf-algorithms} depicts implementation behaviors, with the top panels tracking how placements are sequenced over time in buffer space and the bottom panels showing the resulting distributions of retained time points across history.
We leave formal descriptions of underlying indexing algorithms to future work.
However, reference Python implementations can be found in provided software materials.
For this work, we also translated the tilted algorithm to the general-purpose Zig programming language and then to the Zig-like Cerebras Software Language for use on the WSE.

These algorithms are notable in providing a novel and highly efficient solution to the more general problem of curating dynamic temporal cross-samples from a data stream, and may lend themselves to a broad set of applications outside the scope of phylogeny tracking \citep{moreno2024algorithms}.




\subsection{WSE Architecture and Programming Model}

The Cerebras Wafer-Scale Engine comprises a networked grid of independently executing compute cores (Processing Elements or PEs).
Each PE contains dedicated message-handling infrastructure, which routes tagged 32-bit packets (``wavelets'') to neighboring PEs and/or to be processed locally, according to a programmed rule set.
Each PE contains 48kb of private on-chip memory wideth single clock cycle latency;
communication to neighboring PEs, too, incurs low latency \citep{buitrago2021neocortex}.
PEs support standard arithmetic and flow-control operations, as well as vectorized 32- and 16-bit integer and floating-point operations.

The WSE device is programmed by providing ``kernel'' code, written in the Cerebras Software Language (CSL), to be executed on each PE.
This language's programming model purposefully reflects characteristics of the underlying WSE architecture.
CSL uses an event-driven framework, with tasks defined to respond to wavelet activation signals exchanged between --- and within --- PEs.
Scheduling of active tasks occurs via hardware-level microthreading, which allows for some level of concurrency.
Special faculties are provided for asynchronous send-receive operations that exchange buffered data between PEs.

For further detail, we refer readers to extensive developer documentation made available by Cerebras through their SDK program.
Access can be requested, currently free of charge, via their website.
For this initial work, we evaluated some CSL code on a virtualized $3\times3$ PE array emulated with conventional CPU hardware, while other experiments used hundreds of thousands of PEs on a physical CS-2 device.

\subsection{Asynchronous Island-model Genetic Algorithm}

\input{fig/async-ga-schematic}
\input{fig/benchmarking}

We apply an island-model genetic algorithm, common in applications of parallel and distributed computing to evolutionary computation, to instantiate a wafer-scale evolutionary process spanning PEs \citep{bennett1999building}.
Under this model, PEs host independent populations that interact through migration (i.e., genome exchange) between neighbors.

Our implementation unfolds according to a generational update cycle.
Migration, depicted as blue-red arrows, is handled first.
Each PE maintains independent immigration buffers and emigration buffers dedicated to each cardinal neighbor, depicted by Figure \ref{fig:async-ga-schematic} in solid blue and red, respectively.
On simulation startup, asynchronous receive operations are opened to accept genomes from each neighboring PE into its corresponding immigration buffer.
At startup, additionally, each emigration buffer is populated with genomes copied from the population and an asynchronous send request is opened for each. 
Asynchronous operations are registered to on-completion callbacks that set a per-buffer ``send-'' or ``receive-complete'' flag variable.
In this work, we size send buffers to hold one genome and receive buffers to hold four genomes.
The main population buffer held 32 genomes.

Subsequently, the main update loop tests all completion flags.
For each immigration flag that is set, buffered genomes are copied into the main population buffer, replacing randomly chosen population members.
Then, the flag is reset and a new receive request is initiated.
Likewise, for each emigration flag set, corresponding send buffers are re-populated with randomly sampled genomes from the main population buffer.
Corresponding flags are then reset and new send requests are initiated.
Figure \ref{fig:async-ga-schematic-controlflow} summarizes the interplay between send/receive requests, callbacks, flags, and buffers each generation (``main cycle'').

The remainder of the main update loop handles evolutionary operations within the scope of the executing PE.
Each genome within the population is evaluated to produce a floating point fitness value.
For this initial work, we use a trivial fitness function that explicitly models additive accumulation of beneficial/deleterious mutations as a floating point value within each genome.
After evaluation, tournament selection is applied.
Each slot in the next generation is populated with the highest-fitness genome among $n=5$ randomly sampled individuals, ties broken randomly.

Finally, a mutational operator is applied across all genomes in the next population.
Experiments used a simple Gaussian mutation on each genome's stored fitness value, with sign restrictions used to manipulate the character of selection in some treatments.
At this point, hereditary stratigraphy annotations --- discussed next --- are updated to reflect an elapsed generation.

The next generation cycle is then launched by a self-activating wavelet, repeating until a generation count halting condition is met.

\subsection{Genome Model}

We used a 96-bit genome for clade reconstruction trials, shown in Supplementary Figure \ref{fig:genome-layout} \citep{moreno2024supplement}.
At the outset of simulation, the first 16 bits of founding genomes were randomized and, subsequently, were inherited without mutation, thus identifying descendants of the same founding ancestor.
The next 80 bits were used for hereditary stratigraph annotation, 16 bits for a generation counter and the remaining 64 as a tilted-retention surface with single-bit differentiae.
Other experiments use a 128-bit genome layout, with the first 32 bits used for a floating point ``fitness'' value and the generation counter upgraded from 16 to 32 bits.

\subsection{Software and Data Availability}

Software, configuration files, and executable notebooks for this work are available via Zenodo at \href{https://doi.org/10.5281/zenodo.10974998}{\texttt{doi.org/10.5281/zenodo.10974998}}.
Data and supplemental materials are available via the Open Science Framework at \href{https://osf.io/bfm2z/}{\texttt{osf.io/bfm2z/}} \citep{foster2017open}.

Hereditary stratigraphy utilities are published in the \texttt{hstrat} Python package \citep{moreno2022hstrat}.
This project used data formats from the ALife Standards project \citep{lalejini2019data} and benefited from open-source scientific software \citep{huerta2016ete,2020SciPy-NMeth,harris2020array,reback2020pandas,mckinney-proc-scipy-2010,cock2009biopython,waskom2021seaborn,hunter2007matplotlib,moreno2024apc,moreno2024pecking,moreno2024hsurf,moreno2024wse,dolson2024phylotrackpy}.

WSE experiments reported in this work used the CSL compiler and CS-2 hardware emulator bundled with the Cerebras SDK v1.0.0 \citep{selig2022cerebras}, available via request from Cerebras.
SDK utilities can be run from any Linux desktop environment, regardless of access to Cerebras hardware.
We accessed CS-2 hardware through PSC Neocortex \citep{buitrago2021neocortex}.

%% file: fig/surf-algorithms.tex
\begin{figure*}[h!]
  \centering
\begin{subfigure}[b]{0.43\linewidth}
\centering
\textbf{steady retention policy}
\end{subfigure}
\begin{subfigure}[b]{0.43\linewidth}
\centering
\textbf{tilted retention policy}
\end{subfigure}

\vspace{-1.5ex}

  \begin{subfigure}[b]{0.5\linewidth}
    \centering
  \includegraphics[width=0.88\linewidth,trim={0 0 2.4cm 0},clip]{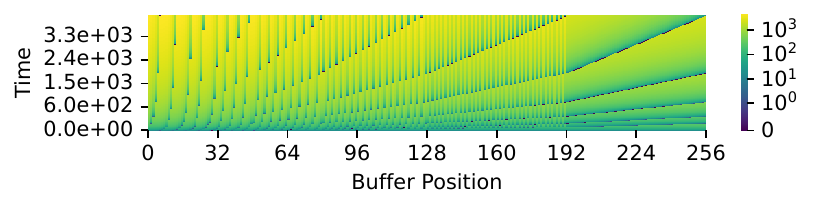}


  \end{subfigure}%
  \begin{subfigure}[b]{0.5\linewidth}
    \centering
  \includegraphics[width=0.88\linewidth, trim={2.4cm 0 0 0},clip]{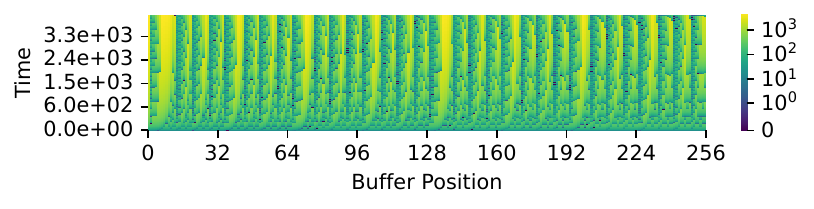}


  \end{subfigure}

\vspace{-1ex}

\begin{subfigure}[b]{0.5\linewidth}
  \flushleft
\includegraphics[height=0.85in, trim={0 0 0 0},clip]{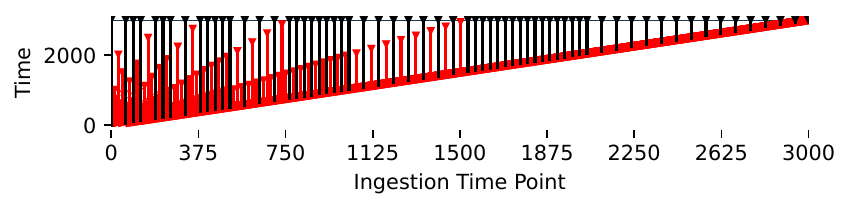}


  \label{fig:steady-retention-plot}
\end{subfigure}%
\begin{subfigure}[b]{0.5\linewidth}
  \centering
\includegraphics[height=0.85in, trim={1.65cm 0 0 0},clip]{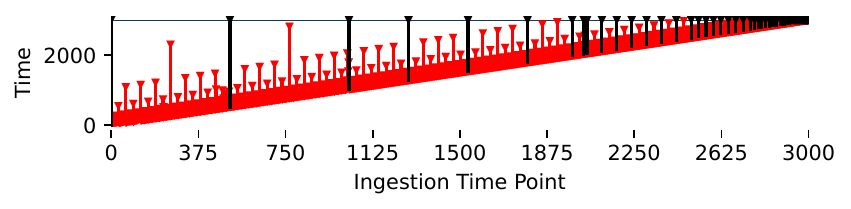}


\end{subfigure}

\vspace{-2.5ex}

\caption{%
  \textbf{Surface-based hereditary stratigraphy implementations.}
  \footnotesize
  Visualizations of steady (left) and tilted (right) surface site selection policies.
  Top-row heatmaps show evolution of time-since-last-deposition for each site on a 256-bit field over the course of 4,096 time steps.
  The bottom row shows retention spans for 3,000 ingested time points.
  Vertical lines span durations between ingestion and elimination for differentia appended at successive time points.
  Time points previously eliminated are marked in red.
  Time elapses from bottom to top in both visualizations.
  }
\label{fig:surf-algorithms}
\vspace{-0.2in}

\end{figure*}

%% file: fig/surf-vs-column-schematic.tex
\begin{figure}[b!]
  \centering
  \includegraphics[width=0.95\linewidth]{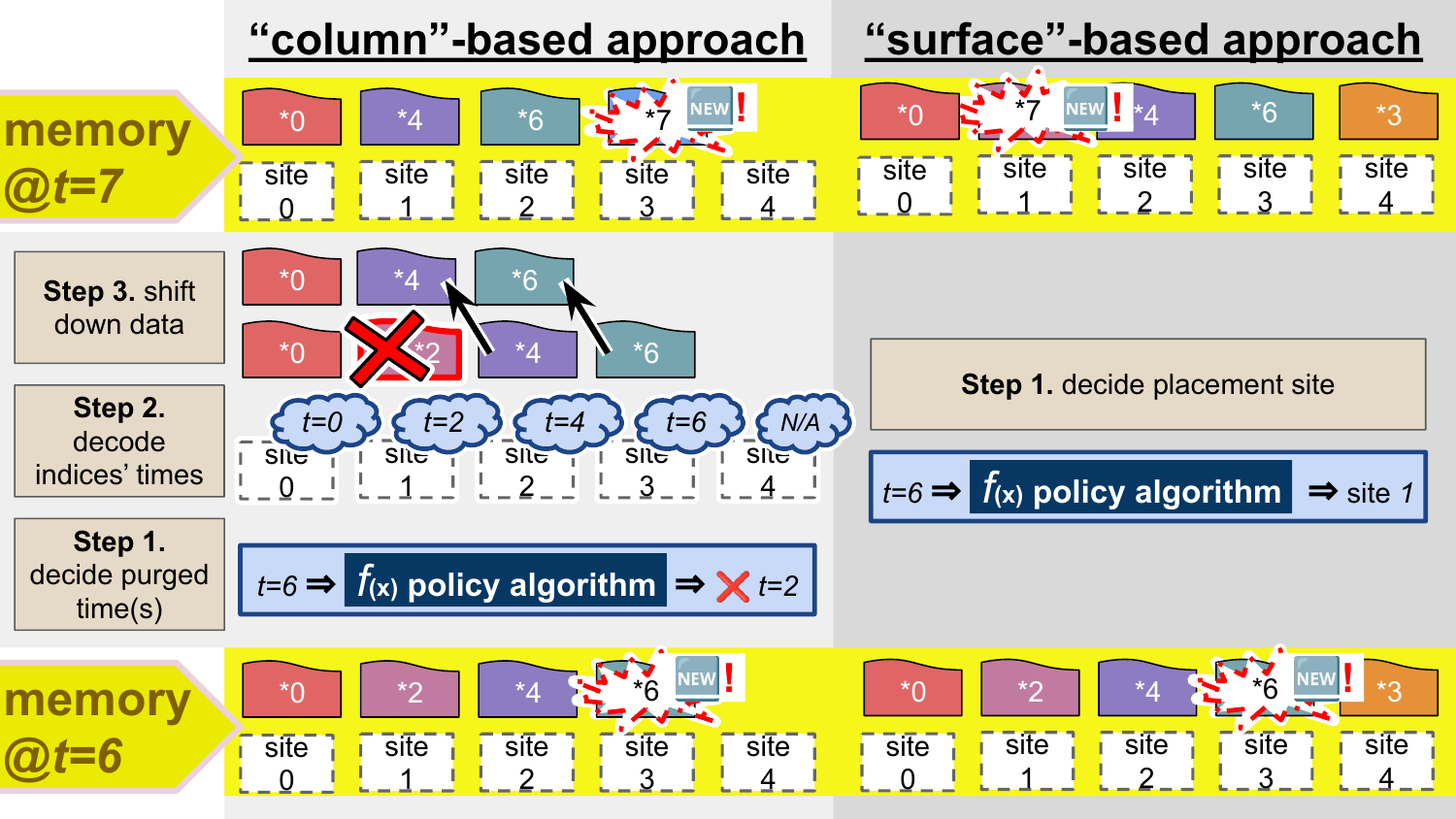}

  \vspace{-1.5ex}

  \caption{
  \textbf{Column vs. surface-based hereditary stratigraphy.}
  \footnotesize
  Contrast of existing sorted-order ``column''-based stratum retention framework with proposed explicitly addressed ``surface''-based approach.}
  \label{fig:surf-vs-column-schematic}
  \vspace{-0.2in}
\end{figure}

%% file: fig/async-ga-schematic.tex
\begin{figure}[t!]
  \centering
  \begin{subfigure}{0.5\linewidth}
    \centering
    \footnotesize
    \includegraphics[width=0.8\linewidth]{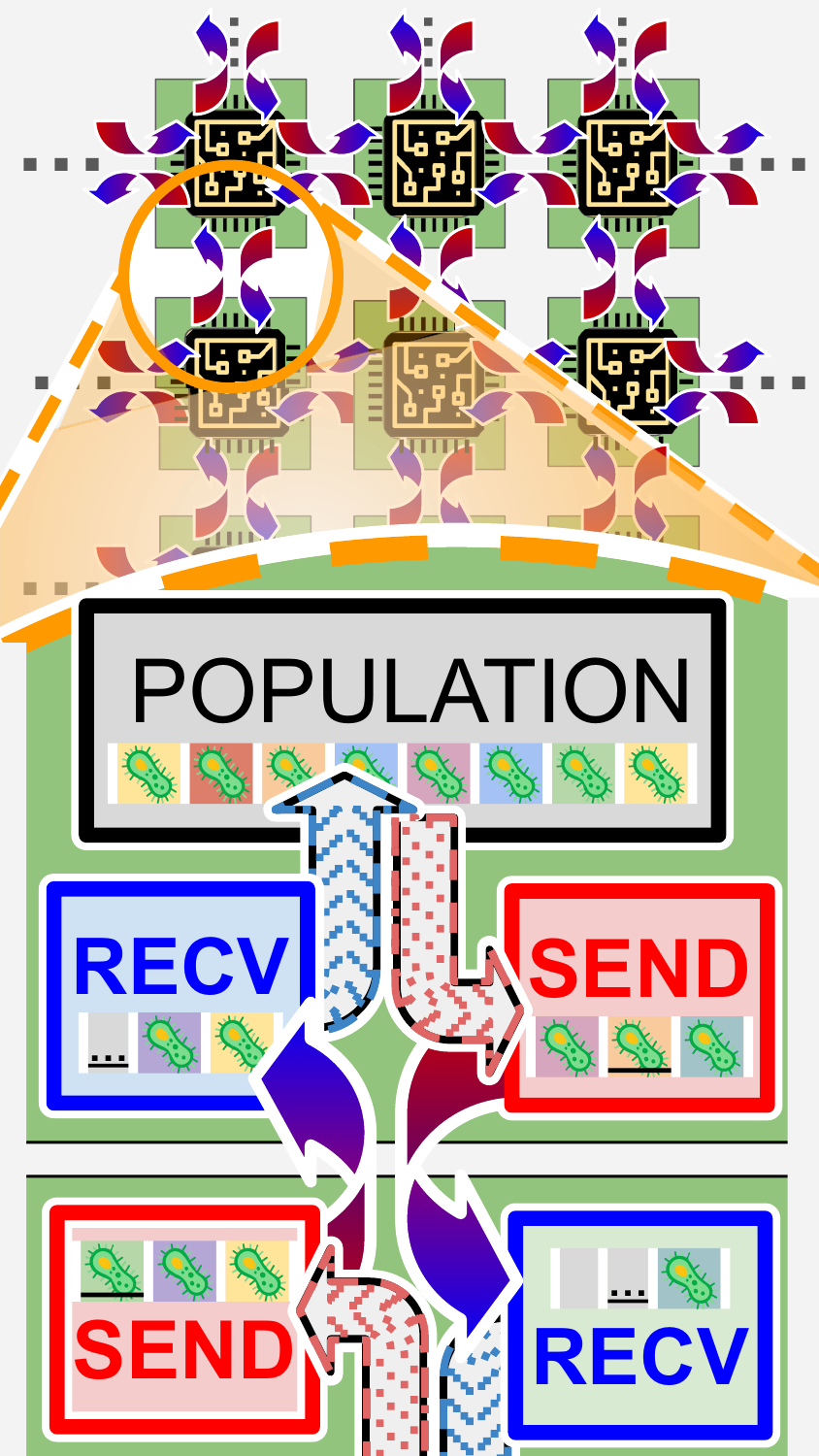}
    \vspace{-0.05in}
    \caption{data flow between PEs}
    \label{fig:async-ga-schematic-dataflow}
  \end{subfigure}%
  \begin{subfigure}{0.5\linewidth}
    \footnotesize
    \centering
    \includegraphics[width=0.8\linewidth]{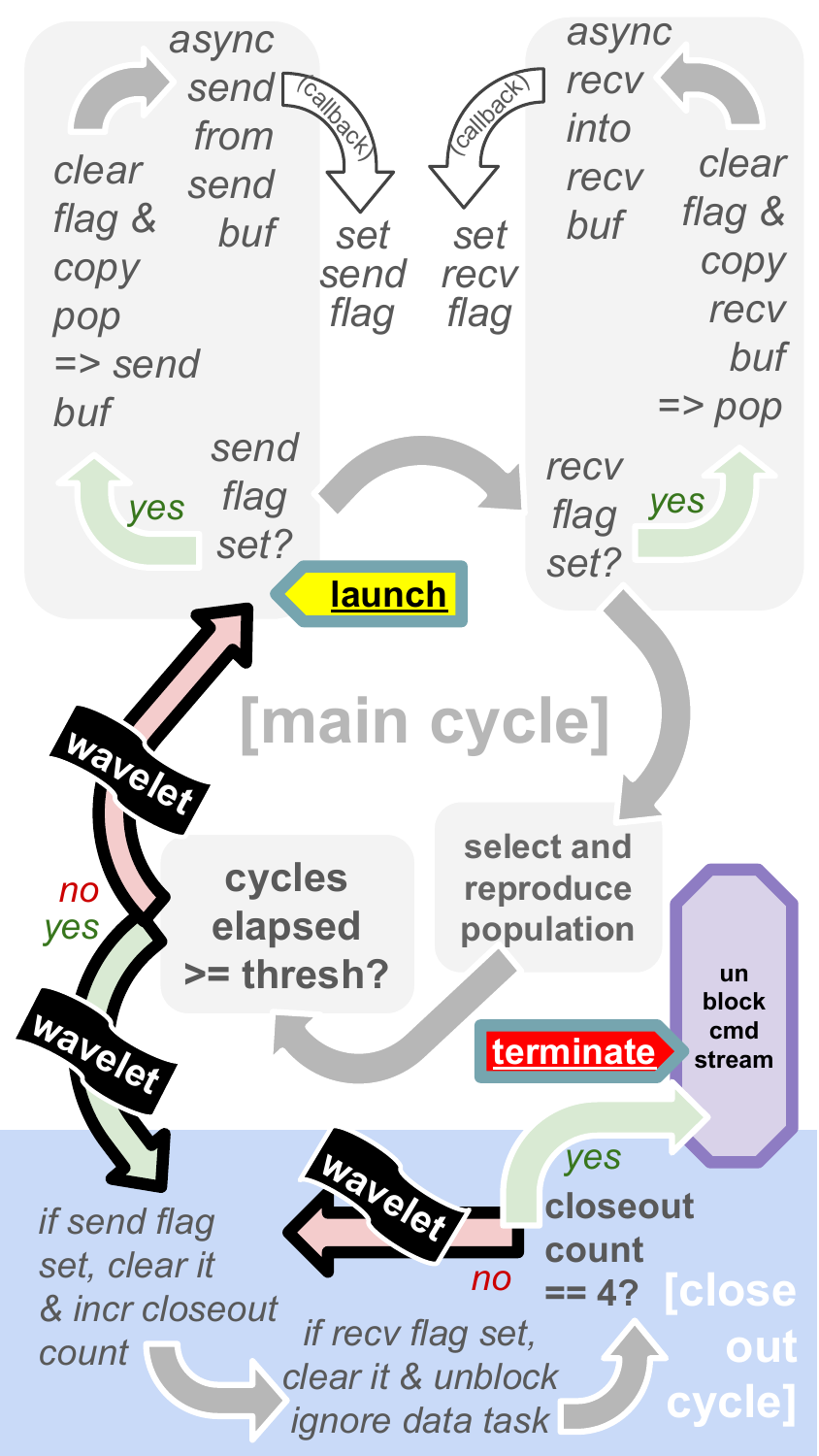}%
    \vspace{-0.05in}
    \caption{control flow within PE}
    \label{fig:async-ga-schematic-controlflow}
  \end{subfigure}
  \vspace{-0.25in}
  \caption{%
  \textbf{Island model GA implementation for WSE.}
  \footnotesize
  Neighboring PEs exchange agents (🦠) via asynchronous send/receive operations from dedicated buffers (``migration''), with on-completion callbacks setting ``ready'' flags to copy between main population and ready buffer.
  }
  \label{fig:async-ga-schematic}
  \vspace{-0.2in}
\end{figure}

%% file: fig/benchmarking.tex
\begin{figure*}[!htb]
  \centering
  \includegraphics[width=0.85\textwidth,trim={0 0 0 0.4in
  }, clip]{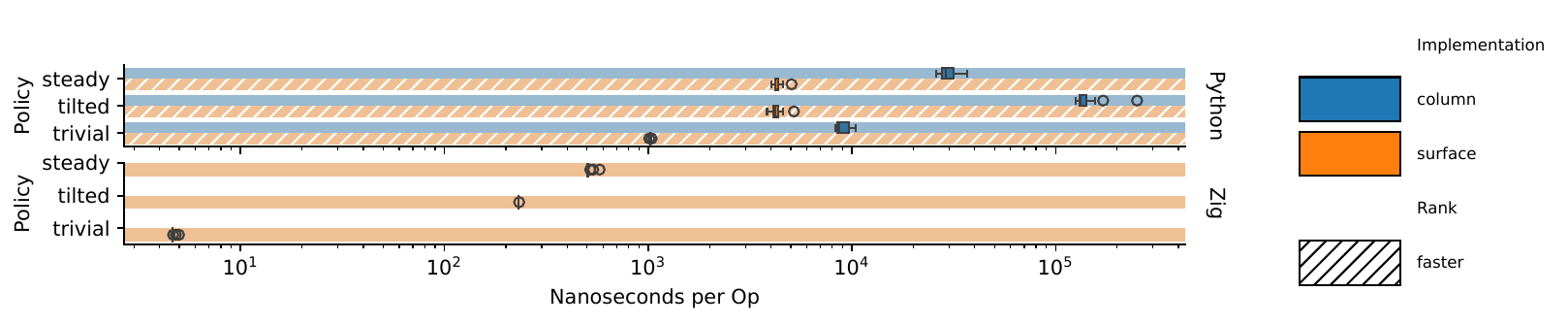}

\vspace{-2.5ex}

  \caption{%
    \textbf{Hereditary stratigraphy algorithm benchmarks.}
    \footnotesize
    Comparison of per-generation operation time for column- and surface-based steady and tilted retention policies, lower is better.
    Top and bottom panels show Python and Zig implementations, respectively.
    Trivial is a simple hardcoded retention decision, provided as a baseline control.
    Background hatching indicates significant outcomes (Mann-Whitney U test; $n=20$).
  }
  \label{fig:benchmarking}
  \vspace{-0.2in}
\end{figure*}

%% file: text/body/results-and-discussion.tex
\section{Results and Discussion} \label{sec:results}

Here, we report a series of benchmarks that evaluate the viability of our proposed approaches to harness the CS-2 accelerator for digital evolution simulations and extract phylogenetic records from said simulations.
First, we compare the performance characteristics of new surface-oriented hereditary stratigraphy methods against preceding column-based implementation to determine the extent to which they succeed in streamlining runtime operations.
Then, we benchmark phylogeny-tracked WSE genetic algorithm implementations.
To assess the runtime overhead of surface-based tracking, we compare against benchmarking results with phylogenetic tracking disabled.
Finally, we validate the credibility of the presented end-to-end annotation-to-reconstruction pipeline by reviewing clade structure and phylometric properties from emulated and on-hardware runs.

\subsection{Surface Algorithm Benchmark on Conventional CPU}

We performed microbenchmark experiments to test computational efficiency of new surface-based algorithms.
Trials measured the real-time speed of sequential generation updates on one annotation with capacity for 64 differentiae.
Benchmarks used both Python, for comparability with existing column algorithm implementations, and Zig, to assess performance under compiler optimization.
Figure \ref{fig:benchmarking} overviews results.

Python implementations of the surface tilted and steady algorithms both took around 4.2 microseconds per operation (standard error [SE] 50ns and 66ns; $n=20$).
For context, this speed was about $4\times$ the measured time for a surface placement using a trivial calculation (SE 0.05; $n=20$).
As expected, column implementations of steady and tilted fared much worse, taking about $7\times$ and $34\times$ the execution time per operation compared to the surface operations.
In both cases, surface implementations significantly outperformed their column counterparts (Mann-Whitney U test, $\alpha = 0.05$).

Zig microbenchmarks clock tilted surface annotation updates at 230 ns per operation (SE 0.9ns; $n=20$).
For context, this time is a little more than twice that required for a main memory access in contemporary computing hardware \citep{markus2022memory}.
Our results measure the operation at $49\times$ the measured time for a trivial placement calculation (SE 0.03; $n=20$).
Zig steady implementation clocks 511 ns per operation (SE 4; $n=20$), $110\times$ trivial (SE 0.8ns; $n=20$).
Note that speedup of Zig implementations relative to Python reflects an intrinsic performance penalty due to interpreter overhead of Python evaluation.
However, comparisons among Python implementations and among Zig implementations are nonetheless informative because all are on equal footing in this regard.


Low-hanging speedups and optimizations exist to further improve the per-operation surface update performance achieved in practice.
Half of update operations on surfaces with single-bit differentiae can be skipped entirely, owing to 50\% probability that randomization fails to change a stored differentia value.
Further, simulations with synchronous or near-synchronous generations can cache calculated surface-placement indices, meaning they would only need to be computed once for an entire subpopulation.
Another possibility is to coarsen temporal resolution, only updating annotations at intervals every $n$th generation.

\subsection{Wafer-Scale Engine Island-model GA Benchmark}

Next, we used an emulator to characterize the expected performance of our island model genetic algorithm on WSE hardware and estimate the magnitude of simulation that might be achieved with on-device execution.
We used the emulator's per-PE clock cycle counters to measure the amount of real time elapsed over the course of a 40-generation-cycle simulation.
We tested using a $3\times3$ PE collective with the tagged 3-word genome and a per-PE population size of 32, applying a tilted hereditary stratigraphy every generation.
PEs completed a mean of 24,138 generation cycles per second (SEM 99; $n=9$).
As an indicator of inter-PE exchange throughput, each PE immigrated a mean of 118 genomes (SEM 11; $n=9$) over 40 elapsed generation cycles.

What scale of simulation does this performance imply at full scale on CS-2 hardware?
Across eight on-device, tracking-enabled trials of 1 million generations (described in the following section), we measured a mean simulation rate of 17,688 generations per second for 562,500 PEs ($750\times750$ rectangle) with run times slightly below one minute.
Trials with 1,600 PEs ($40\times40$) performed similarly, completing 17,734 generations per second.

Multiplied out to a full day, 17,000 generations per second turnover would elapse around 1.5 billion generations.
With 32 individuals hosted per each of 850,000 PEs, the net population size would sit around 27 million at full CS-2 wafer scale.
(Note, though, that available on-chip memory could support thousands of our very simple agents per PE, raising the potential for a net population size on the order of a billion agents.)
A naive extrapolation estimates on the order of a quadrillion agent replications could be achieved hourly at full wafer scale for such a very simple model.
We look forward to conducting more thorough benchmarking and scaling experiments in future work.

How fast is simulation without hstrat instrumentation?
We repeated our hardware-emulated benchmark with 32-bit genomes stripped of all instrumentation.
Under these conditions, PEs completed on average 47,175 generations per second (SEM 220; $n=9$) and immigrated 118 genomes (SEM 12; $n=9$).

These timings measure phylogeny tracking as approximately equivalent to that of the other aspects of simulation, combined.
Given the highly minimalistic nature of the agent model and selection process, this result is highly promising.
In actual use, most experiments will likely involve a much more sophisticated agent model, so relative overhead of tracking will be diminished.
Additionally, these results do not include caching and coarsening strategies discussed above, which would speed tracking up considerably.

\subsection{Clade Reconstruction Trial}

To assess overall correctness of our Cerebras Software Language (CSL) surface algorithm implementation, we reviewed the clade structure of sample phylogenetic reconstructions from emulated WSE hardware.
Full reconstruction quality testing of the underlying new surface algorithms themselves is provided in other recent work \citep{moreno2024guide}.
(There, we found quality to be comparable to existing column algorithms.)
For these experiments, we tagged genomes with a 16-bit randomly generated identifier at simulation startup, as shown in Supplementary Figure \ref{fig:genome-layout} \citep{moreno2024supplement}.
We instantiated populations over a $3\times3$ PE collective for durations of 25, 50, 100, and 250 generation cycles with neutral selection.
Phylogenetic reconstruction used one end-state genome sampled per PE.

\input{fig/tagged.tex}

Figure \ref{fig:tagged} shows reconstructed phylogenies from each duration.
As expected, taxa belonging to the same founding clade (shown by color) are reconstructed as more closely related to each other than to other taxa.
Additionally, and also as expected, the number of distinct remaining founding clades diminishes monotonically with increasing simulation duration.
These consistencies corroborate the general integrity of our CSL surface implementation.
Note, however, the incidence of moderate overestimations for relatedness between independently-tagged clades throughout.
As mentioned earlier, and discussed in greater depth elsewhere \citep{moreno2024guide}, this is an expected artifact of hereditary stratigraphy with single-bit differentiae.
Applications requiring greater reconstruction precision can opt for larger differentia sizes and/or higher differentia counts.

\subsection{On-hardware Trial}

\input{fig/on-device}

Finally, we set out to assess the performance of our pipeline at full wafer scale.
For these experiments, we used the four-word, tracking-enabled genome layout, with the full first 32 bits containing a floating point fitness value.
We defined two treatments: \textit{\textbf{purifying-only conditions}}, where 33\% of agent replication events decreased fitness by a normally-distributed amount, and \textit{\textbf{adaption-enabled conditions}}, which added beneficial mutations that increased fitness by a normally-distributed amount, occurring with 0.3\% probability.
These beneficial mutations introduced the possibility for selective sweeps.
As before, we used tournament size 5 for both treatments.
We performed four on-hardware replicates of each treatment instantiated on 10k ($100\times100$), 250k ($500\times500$) and 562.5k ($750\times750$) PE arrays.
We halted each PE after it elapsed 1 million generations.

Upon completion, we sampled one genome from each PE.
Then, we performed an agglomerative trie-based reconstruction from subsamples of 10k end-state genomes \citep{moreno2024analysis}.
Figure \ref{fig:on-device} compares phylogenies generated under the purifying-only and adaption-enabled treatments.
As expected \citep{moreno2023toward}, purifying-only treatment phylogenies consistently exhibited greater sum branch length and mean evolutionary distinctiveness, with the effect on branch length particularly strong.
These structural effects are apparent in example phylogeny trees from 562.5k PE trials (Figures \labelcref{fig:on-device-adaptive,fig:on-device-purifying}).
Successful differentiation between treatments is a highly promising outcome.
This result not only supports the correctness of our methods and implementation, but also confirms the capability of reconstruction-based analyses to meaningfully describe dynamics of very large-scale evolution simulations.

%% file: fig/tagged.tex
\begin{figure}

\begin{subfigure}[c]{\linewidth} \centering
\begin{minipage}[c]{0.08\linewidth} \flushright
    \caption{\rotatebox[origin=c]{90}{25 cycles}}
    \label{fig:tagged_25}
  \end{minipage}%
  \begin{minipage}[c]{0.92\linewidth}
    \includegraphics[width=\textwidth,height=0.6in,trim={0 0.81cm 0 0},clip]{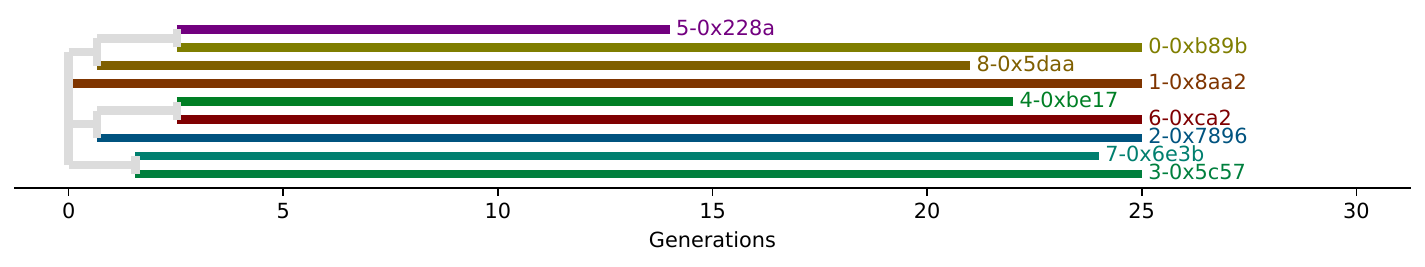}
  \end{minipage}%
\end{subfigure}

\vspace{-1ex}

\begin{subfigure}[c]{\linewidth} \centering
\begin{minipage}[c]{0.08\linewidth} \flushright
    \caption{\rotatebox[origin=c]{90}{50 cycles}}
    \label{fig:tagged_50}
  \end{minipage}%
  \begin{minipage}[c]{0.92\linewidth}
    \includegraphics[width=\textwidth,height=0.6in,trim={0 0.81cm 0 0},clip]{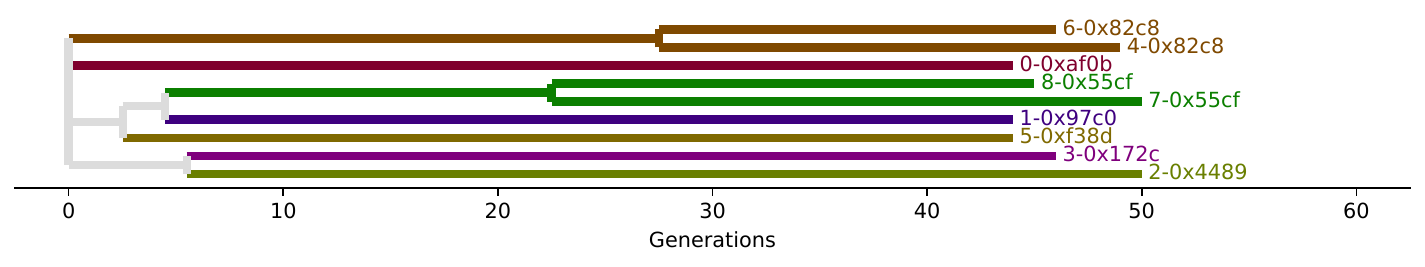}
  \end{minipage}%
\end{subfigure}

\vspace{-1ex}

\begin{subfigure}[c]{\linewidth} \centering
  \begin{minipage}[c]{0.08\linewidth} \flushright
    \caption{\rotatebox[origin=c]{90}{100 cycles}}
    \label{fig:tagged_100}
  \end{minipage}%
  \begin{minipage}[c]{0.92\linewidth}
    \includegraphics[width=\textwidth,height=0.6in,trim={0 0.81cm 0 0},clip]{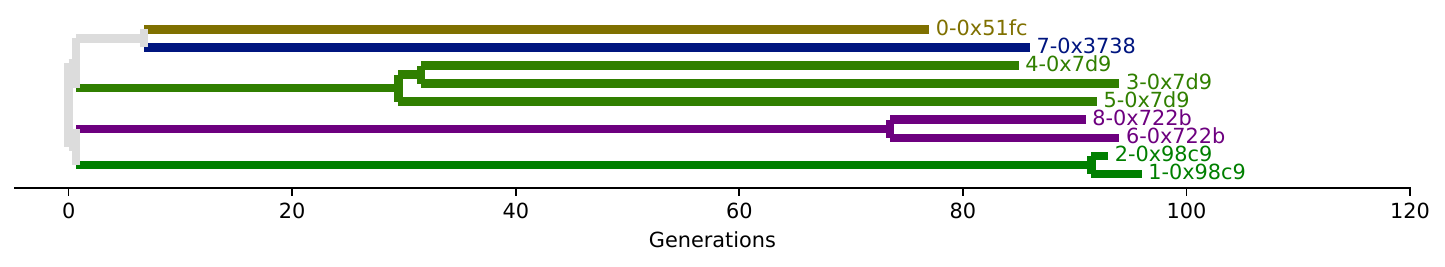}
  \end{minipage}%
\end{subfigure}

\vspace{-1ex}

\begin{subfigure}[c]{\linewidth} \centering
  \begin{minipage}[c]{0.08\linewidth} \flushright
    \caption{\rotatebox[origin=c]{90}{250 cycles}}
    \label{fig:tagged_250}
  \end{minipage}%
  \begin{minipage}[c]{0.92\linewidth}
    \includegraphics[width=\textwidth,height=0.8in]{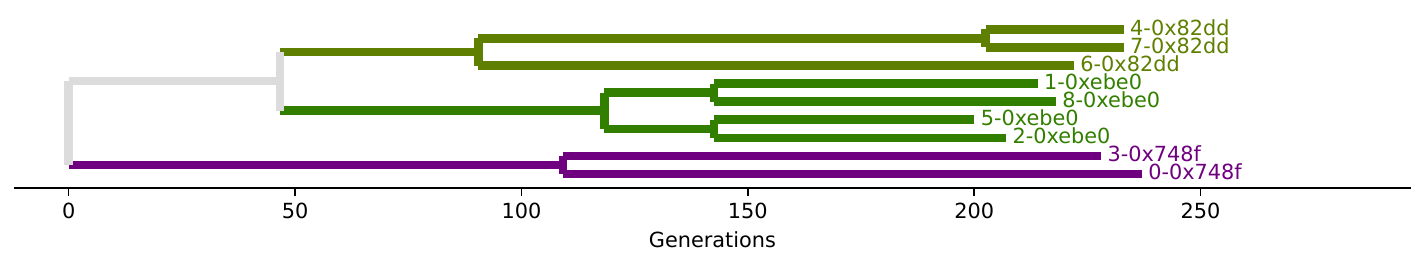}
  \end{minipage}%
\end{subfigure}

\vspace{-2ex}

\caption{%
\textbf{Clade Reconstruction Trial.}
\footnotesize
Example phylogenies reconstructed from runs of increasing duration on a virtual grid of nine hardware-simulated PEs.
Founding genomes were tagged with random 16-byte identifier values, which were held constant throughout simulation (Supplementary Figure \ref{fig:genome-layout}).
Color-coding indicates each sampled taxon's founding ancestor according to this identifier value.
Simulation was performed with neutral selection.
}
\label{fig:tagged}
\vspace{-0.2in}
\end{figure}

%% file: fig/on-device.tex
\begin{figure}

\begin{subfigure}[c]{\linewidth}
  \centering
    \includegraphics[width=0.9\linewidth]{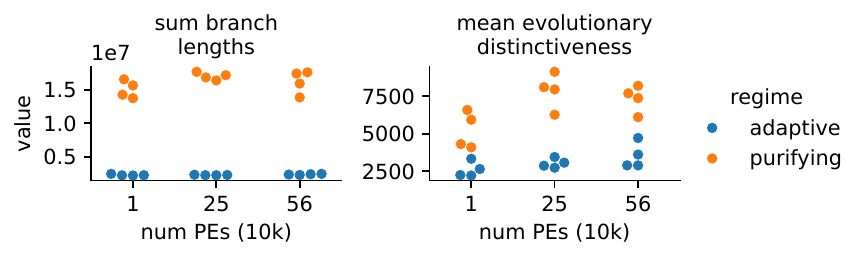}
    \vspace{-1.8ex}
    \caption{\footnotesize adaptive vs. purifying phylometric structure}
  \label{fig:on-device-phylometrics}
\end{subfigure}

\vspace{0.8ex}

\begin{subfigure}[c]{0.5\linewidth}
  \centering
  \includegraphics[width=0.7\linewidth,trim={0 11.05in 7.2in 0in},clip]{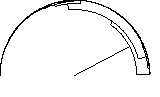}

  \vspace{-2ex}
  \includegraphics[width=0.7\linewidth,trim={0 11.05in 7.2in 0in},clip]{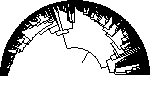}

  \vspace{-3ex}
  \footnotesize
 \caption{\footnotesize adaptive regime}
  \label{fig:on-device-adaptive}
\end{subfigure}%
\begin{subfigure}[c]{0.5\linewidth}
  \centering
  \includegraphics[width=0.7\linewidth,trim={0 11.05in 7.2in 0in},clip]{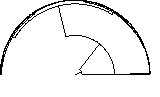}

  \vspace{-2ex}
  \includegraphics[width=0.7\linewidth,trim={0 11.05in 7.2in 0in},clip]{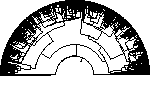}

  \vspace{-3ex}
  \footnotesize
 \caption{\footnotesize purifying regime}
  \label{fig:on-device-purifying}
\end{subfigure}

\vspace{-1.5ex}

\caption{%
\textbf{On-hardware Trial.}
\footnotesize
Results from phylogenetic reconstruction of 1 million generation on-hardware simulations.
Panel \ref{fig:on-device-phylometrics} compares phylometric readings from purifying-only and adaptation-enabled configurations, 4 replicates each.
Panels \labelcref{fig:on-device-adaptive,fig:on-device-purifying} juxtapose example $750\times750$ PE simulation phylogenies under each configuration regime.
Phylometrics were calculated from reconstructions with 10k sampled end-state agents.
For legibility, phylogeny visualizations were further subsampled to 1k end-state agents.
Top phylogenies are linear-scaled.
Bottom phylogenies are log-scaled with ultrametric correction to better show topology.
}
\label{fig:on-device}
\vspace{-0.2in}
\end{figure}

%% file: text/body/conclusion.tex
\section{Conclusion} \label{sec:conclusion}

Computing hardware with transformative capabilities is presently coming to market.
This fact presents an immediate opportunity to bring orders-of-magnitude greater simulation scale to bear on grand challenges in artificial life.
It is not unreasonable to anticipate the possibility that with such resources some aspects of these open questions will be revealed to harbor more-is-different dispositions, in which larger scales reveal qualitatively different dynamics \citep{anderson1972more}.
Riding the coattails of AI-workload-driven hardware development, itself largely driven by profound more-is-different payoffs in deep learning, provides perhaps the most immediate means toward this possibility.

Such an endeavor is a community-level challenge that will require significant resources and collaborative effort.
Work presented here is an early step in methods and infrastructure development necessary to scale up what is possible in digital evolution research.
We have demonstrated new algorithms and software for phylogeny-enabled agent-based evolution on a next-generation HPC accelerator hardware platform.
Microbenchmarking results show that proposed instrumentation algorithms achieve several-fold improvement in computational efficiency.
Related work shows new algorithms to improve reconstruction quality in some cases, too \citep{moreno2024guide}.
Benchmarks confirm that, including tracking operations, simple models at wafer scale can achieve quintillions of replications per day.
In future work, it will be necessary to move beyond proof-of-concept and explore the limits of these capabilities in the context of more demanding, interaction-intensive models.

Special characteristics set agent-based digital evolution apart from many other HPC application domains and position it as a potentially valuable testbed for innovation and leadership.
Among these factors are challenging workload heterogeneity (varying within a population and over evolutionary time), resiliency of global state to local perturbations, and perhaps substantial freedom to recompose underlying simulation semantics to accommodate hardware capabilities.
Indeed, artificial life and digital evolution have played host to notable boundary-pushing approaches to computing at the frontiers of computing modalities such as best-effort computing, reservoir computing, global-scale time-available computing, and modular tile-based computing in addition to more traditional cluster and GPU-oriented approaches \citep{moreno2021conduit,ackley2020best,ackley2023robust,heinemann2008artificial,miikkulainen2024evolving,ray1995proposal}.
Work done to scale up digital evolution simulation should be done with an eye for contributions back to broader HPC constituencies.

In this vein, presented ``surface'' indexing algorithms stand to benefit larger classes of stream curation problems, situations in which a rolling feed of sequenced observations must be dynamically downsampled to ensure retention of elements representative across observed history \citep{moreno2024algorithms}.
In particular, to further benefit observable agent-based modeling, we are interested in exploring applications that sample time-series activity at simulation sites or distill coarsened agent histories (e.g., position over time).

Our goal in this work is to build new capabilities that empower research agendas across the digital evolution and artificial life community.
To this end, we have prioritized making our CSL and Python software easily reusable by other researchers.
In particular, CSL code implementing the presented island-model GA is modularized and extensible for drop-in customization to instantiate any fixed-length genome content and fitness criteria.
We look forward to collaboration in broader tandem efforts to harness the Cerebras platform, and other emerging hardware, in follow-on work.

%% file: text/acknowledgement.tex
\section*{Acknowledgement}
{\footnotesize
Computational resources were provided by the MSU Institute for Cyber-Enabled Research and the ByteBoost training program.
This research was supported in part the Eric and Wendy Schmidt AI in Science Postdoctoral Fellowship and NSF funding (DEB 1813069).
Any opinions, findings, and conclusions or recommendations expressed in this material are those of the author(s) and do not necessarily reflect the views of the National Science Foundation.
Thank you also to Mathias Jacquelin, Udai Mody, and Leighton Wilson at Cerebras Systems.
}

%% file: text/references.tex
\putbib

%% file: text/supplement.tex
\section{Supplemental Material}

\input{fig/genome-layout}

%% file: fig/genome-layout.tex
\definecolor{LighterBlue}{rgb}{0.84, 0.92, 0.95}
\definecolor{LighterSalmon}{rgb}{1.0, 0.81, 0.76}
\definecolor{LighterPastelGreenYellow}{rgb}{0.88, 0.96, 0.90}

\newcolumntype{q}{!{\color{white} \vrule width 2pt}}

\begin{figure*}
    \centering
\begin{minipage}[t]{\textwidth}
\centering
\footnotesize
\begin{tabular}{
c
>{\columncolor{LighterBlue}}c
>{\columncolor{LighterBlue}}c
>{\columncolor{LighterSalmon}}c
>{\columncolor{LighterSalmon}}c
q 
>{\columncolor{LighterPastelGreenYellow}}c
>{\columncolor{LighterPastelGreenYellow}}c
>{\columncolor{LighterPastelGreenYellow}}c
>{\columncolor{LighterPastelGreenYellow}}c
q 
>{\columncolor{LighterPastelGreenYellow}}c
>{\columncolor{LighterPastelGreenYellow}}c
>{\columncolor{LighterPastelGreenYellow}}c
>{\columncolor{LighterPastelGreenYellow}}c
}
& \multicolumn{4}{cq}{\cellcolor{white}Word 0} & \multicolumn{4}{cq}{\cellcolor{white}Word 1} & \multicolumn{4}{c}{\cellcolor{white}Word 2} \\
\cmidrule(l{1.5pt}r{1.5pt}){2-5}
\cmidrule(l{1.5pt}r{1.5pt}){6-9}
\cmidrule(l{1.5pt}r{1.5pt}){10-13}
Byte & {\cellcolor{white}0} & {\cellcolor{white}1} & {\cellcolor{white}2} & {\cellcolor{white}3} & {\cellcolor{white}4} & {\cellcolor{white}5} & {\cellcolor{white}6} & {\cellcolor{white}7} & {\cellcolor{white}8} & {\cellcolor{white}9} & {\cellcolor{white}10} & {\cellcolor{white}11} \\
\cmidrule(l{1.5pt}r{1.5pt}){2-2}
\cmidrule(l{1.5pt}r{1.5pt}){3-3}
\cmidrule(l{1.5pt}r{1.5pt}){4-4}
\cmidrule(l{1.5pt}r{1.5pt}){5-5}
\cmidrule(l{1.5pt}r{1.5pt}){6-6}
\cmidrule(l{1.5pt}r{1.5pt}){7-7}
\cmidrule(l{1.5pt}r{1.5pt}){8-8}
\cmidrule(l{1.5pt}r{1.5pt}){9-9}
\cmidrule(l{1.5pt}r{1.5pt}){10-10}
\cmidrule(l{1.5pt}r{1.5pt}){11-11}
\cmidrule(l{1.5pt}r{1.5pt}){12-12}
\cmidrule(l{1.5pt}r{1.5pt}){13-13}
& \multicolumn{4}{cq}{\cellcolor{white}} & \multicolumn{4}{cq}{\cellcolor{white}} & \multicolumn{4}{c}{\cellcolor{white}} \\[-2ex]
\scriptsize{Genome 0} & \texttt{F9} & \texttt{02} & \texttt{79} & \texttt{00} & \texttt{8D} & \texttt{22} & \texttt{4F} & \texttt{F3} & \texttt{D2} & \texttt{78} & \texttt{AD} & \texttt{C7} \\
& \multicolumn{4}{cq}{\cellcolor{white}} & \multicolumn{4}{cq}{\cellcolor{white}} & \multicolumn{4}{c}{\cellcolor{white}} \\[-2ex]
\scriptsize{Genome 1} & \texttt{F9} & \texttt{02} & \texttt{75} & \texttt{00} & \texttt{8D} & \texttt{A1} & \texttt{CB} & \texttt{F2} & \texttt{D1} & \texttt{5B} & \texttt{CC} & \texttt{D4} \\
& \multicolumn{4}{cq}{\cellcolor{white}} & \multicolumn{4}{cq}{\cellcolor{white}} & \multicolumn{4}{c}{\cellcolor{white}} \\[-2ex]
\scriptsize{Genome 2} & \texttt{61} & \texttt{B6} & \texttt{65} & \texttt{00} & \texttt{66} & \texttt{29} & \texttt{B4} & \texttt{F0} & \texttt{62} & \texttt{99} & \texttt{5A} & \texttt{61} \\
{\cellcolor{white}\ldots} & {\cellcolor{white}\ldots} & {\cellcolor{white}\ldots} & {\cellcolor{white}\ldots} & {\cellcolor{white}\ldots} & {\cellcolor{white}\ldots} & {\cellcolor{white}\ldots} & {\cellcolor{white}\ldots} & {\cellcolor{white}\ldots} & {\cellcolor{white}\ldots} & {\cellcolor{white}\ldots} &
{\cellcolor{white}\ldots} & {\cellcolor{white}\ldots} \\
\end{tabular}
\end{minipage}

\begin{minipage}[t]{\textwidth}
\caption{%
\textbf{Example genomes sampled after validation experiment completion.}
  In validation testing, genomes were composed of three 32-bit words.
  The first two bytes (blue) are fixed random markers generated at simulation start-up, indicating independent lineage originations.
  The next two bytes (salmon) are a generation counter.
  Bits within the final eight bytes are lineage checkpoint values to facilitate phylogenetic reconstruction, arranged according to a tilted hereditary stratigraphic algorithm.
  Note that this genome does not include any content affecting agent traits or fitness --- neutral selection was used for this validation experiment.
}
\label{fig:genome-layout}
\end{minipage}
\end{figure*}

%% file: main.bbl
\begin{thebibliography}{}

\bibitem[Abdelhafez et~al., 2019]{abdelhafez2019performance}
Abdelhafez, A., Alba, E., \& Luque, G. (2019).
\newblock Performance analysis of synchronous and asynchronous distributed
  genetic algorithms on multiprocessors.
\newblock {\em Swarm and Evolutionary Computation}, 49, 147--157.

\bibitem[Ackley, 2011]{livingcomputationSFBSanta}
Ackley, D. (2011).
\newblock {\em {S}{F}{B}: {T}he {S}anta {F}e {B}oard {P}rogrammer\&apos;s
  {R}eference {M}anual --- livingcomputation.com}.
\newblock \url{https://livingcomputation.com/s/doc/}.
\newblock [Accessed 08-04-2024].

\bibitem[Ackley, 2020]{ackley2020best}
Ackley, D.~H. (2020).
\newblock Best-effort computing with spots and spatial threads.
\newblock {\em Artificial Life Conference Proceedings 32}, 13--15.

\bibitem[Ackley, 2023]{ackley2023robust}
Ackley, D.~H. (2023).
\newblock A robust programmable replicator for an indefinitely scalable
  machine.
\newblock {\em ALIFE 2023: Ghost in the Machine: Proceedings of the 2023
  Artificial Life Conference}.

\bibitem[Ackley \& Ackley, 2016]{ackley2016ulam}
Ackley, D.~H. \& Ackley, E.~S. (2016).
\newblock The ulam programming language for artificial life.
\newblock {\em Artificial Life}, 22(4), 431--450.

\bibitem[Ackley et~al., 2013]{ackley2013movable}
Ackley, D.~H., Cannon, D.~C., \& Williams, L.~R. (2013).
\newblock A movable architecture for robust spatial computing.
\newblock {\em The Computer Journal}, 56(12), 1450--1468.

\bibitem[Ackley \& Williams, 2011]{ackley2011homeostatic}
Ackley, D.~H. \& Williams, L.~R. (2011).
\newblock Homeostatic architectures for robust spatial computing.
\newblock {\em 2011 Fifth IEEE Conference on Self-Adaptive and Self-Organizing
  Systems Workshops}, 91--96.

\bibitem[Anderson, 1972]{anderson1972more}
Anderson, P.~W. (1972).
\newblock More is different: Broken symmetry and the nature of the hierarchical
  structure of science.
\newblock {\em Science}, 177(4047), 393--396.

\bibitem[Bennett~III et~al., 1999]{bennett1999building}
Bennett~III, F.~H., Koza, J.~R., Shipman, J., \& Stiffelman, O. (1999).
\newblock Building a parallel computer system for \$18,000 that performs a half
  peta-flop per day.
\newblock {\em Proceedings of the 1st Annual Conference on Genetic and
  Evolutionary Computation-Volume 2}, 1484--1490.

\bibitem[Bohm et~al., 2017]{bohm2017mabe}
Bohm, C., G., N.~C., \& Hintze, A. (2017).
\newblock {MABE} (modular agent based evolver): A framework for digital
  evolution research.
\newblock volume~14 of {\em ECAl the European Conference on Artificial Life},
  76--83.
\newblock \url{https://doi.org/10.7551/ecal_a_016}

\bibitem[Buitrago \& Nystrom, 2021]{buitrago2021neocortex}
Buitrago, P.~A. \& Nystrom, N.~A. (2021).
\newblock Neocortex and bridges-2: A high performance ai+ hpc ecosystem for
  science, discovery, and societal good.
\newblock {\em High Performance Computing: 7th Latin American Conference, CARLA
  2020, Cuenca, Ecuador, September 2--4, 2020, Revised Selected Papers 7},
  205--219.

\bibitem[Burke et~al., 2003]{burke2003increased}
Burke, E., Gustafson, S., Kendall, G., \& Krasnogor, N. (2003).
\newblock Is increased diversity in genetic programming beneficial? an analysis
  of lineage selection.
\newblock {\em The 2003 Congress on Evolutionary Computation, 2003. CEC '03.},
  volume~2, 1398--1405 Vol.2.
\newblock \url{https://doi.org/10.1109/cec.2003.1299834}

\bibitem[Cant{\'u}-Paz, 2001]{cantu2001master}
Cant{\'u}-Paz, E. (2001).
\newblock Master-slave parallel genetic algorithms.
\newblock {\em Efficient and Accurate Parallel Genetic Algorithms}, 33--48.
  Springer.

\bibitem[Cock et~al., 2009]{cock2009biopython}
Cock, P.~J., Antao, T., Chang, J.~T., Chapman, B.~A., Cox, C.~J., Dalke, A.,
  Friedberg, I., Hamelryck, T., Kauff, F., Wilczynski, B., et~al. (2009).
\newblock Biopython: freely available python tools for computational molecular
  biology and bioinformatics.
\newblock {\em Bioinformatics}, 25(11), 1422--1423.

\bibitem[Cohen, 1987]{cohen1987computer}
Cohen, F. (1987).
\newblock Computer viruses: theory and experiments.
\newblock {\em Computers \& security}, 6(1), 22--35.

\bibitem[De~Rainville et~al., 2012]{de2012deap}
De~Rainville, F.-M., Fortin, F.-A., Gardner, M.-A., Parizeau, M., \& Gagn\'{e},
  C. (2012).
\newblock {DEAP}: A {Python} framework for evolutionary algorithms.
\newblock {\em Proceedings of the 14th Annual Conference Companion on Genetic
  and Evolutionary Computation}, Gecco '12, 85–92.
\newblock \url{https://doi.org/10.1145/2330784.2330799}

\bibitem[Dolson et~al., 2024]{dolson2024phylotrackpy}
Dolson, E., Moreno, M.~A., \& rodsan0 (2024).
\newblock {\em emilydolson/phylotrackpy: v0.2.0}.
\newblock \url{https://doi.org/10.5281/zenodo.10888780}

\bibitem[Dolson \& Ofria, 2021]{dolson2021digital}
Dolson, E. \& Ofria, C. (2021).
\newblock Digital evolution for ecology research: a review.
\newblock {\em Frontiers in Ecology and Evolution}, 9, 750779.

\bibitem[Emani et~al., 2021]{emani2021accelerating}
Emani, M., Vishwanath, V., Adams, C., Papka, M.~E., Stevens, R., Florescu, L.,
  Jairath, S., Liu, W., Nama, T., \& Sujeeth, A. (2021).
\newblock Accelerating scientific applications with sambanova reconfigurable
  dataflow architecture.
\newblock {\em Computing in Science \& Engineering}, 23(2), 114--119.

\bibitem[Faith, 1992]{faithConservationEvaluationPhylogenetic1992}
Faith, D.~P. (1992).
\newblock Conservation evaluation and phylogenetic diversity.
\newblock {\em Biological Conservation}, 61(1), 1--10.
\newblock \url{https://doi.org/10.1016/0006-3207(92)91201-3}

\bibitem[Foster \& Deardorff, 2017]{foster2017open}
Foster, E.~D. \& Deardorff, A. (2017).
\newblock Open science framework (osf).
\newblock {\em Journal of the Medical Library Association}, 105(2), 203.

\bibitem[French et~al., 2023]{frenchHostPhylogenyShapes2023}
French, R.~K., Anderson, S.~H., Cain, K.~E., Greene, T.~C., Minor, M.,
  Miskelly, C.~M., Montoya, J.~M., Wille, M., Muller, C.~G., Taylor, M.~W.,
  Digby, A., \& Holmes, E.~C. (2023).
\newblock Host phylogeny shapes viral transmission networks in an island
  ecosystem.
\newblock {\em Nature Ecology \& Evolution}, 1--10.
\newblock \url{https://doi.org/10.1038/s41559-023-02192-9}.
\newblock Publisher: Nature Publishing Group

\bibitem[Friggeri et~al., 2014]{friggeri2014rumor}
Friggeri, A., Adamic, L., Eckles, D., \& Cheng, J. (2014).
\newblock Rumor cascades.
\newblock {\em Proceedings of the International AAAI Conference on Web and
  Social Media}, 8(1), 101--110.
\newblock \url{https://doi.org/10.1609/icwsm.v8i1.14559}

\bibitem[Garwood et~al., 2019]{garwood2019revosim}
Garwood, R.~J., Spencer, A.~R., \& Sutton, M.~D. (2019).
\newblock Revosim: Organism-level simulation of macro- and microevolution.
\newblock {\em Palaeontology}, 62(3), 339--355.
\newblock \url{https://doi.org/10.1111/pala.12420}

\bibitem[Giardina et~al., 2017]{giardina2017inference}
Giardina, F., Romero-Severson, E.~O., Albert, J., Britton, T., \& Leitner, T.
  (2017).
\newblock Inference of transmission network structure from hiv phylogenetic
  trees.
\newblock {\em PLoS computational biology}, 13(1), e1005316.

\bibitem[Godin-Dubois et~al., 2019]{godin2019apoget}
Godin-Dubois, K., Cussat-Blanc, S., \& Duthen, Y. (2019).
\newblock Apoget: Automated phylogeny over geological timescales.
\newblock {\em ALIFE 2019 (MethAL workshop)}.
\newblock \url{https://doi.org/10.13140/rg.2.2.33781.93921}

\bibitem[Good et~al., 2017]{good2017dynamics}
Good, B.~H., McDonald, M.~J., Barrick, J.~E., Lenski, R.~E., \& Desai, M.~M.
  (2017).
\newblock The dynamics of molecular evolution over 60,000 generations.
\newblock {\em Nature}, 551(7678), 45--50.

\bibitem[Han et~al., 2005]{han2005stream}
Han, J., Chen, Y., Dong, G., Pei, J., Wah, B.~W., Wang, J., \& Cai, Y.~D.
  (2005).
\newblock Stream cube: An architecture for multi-dimensional analysis of data
  streams.
\newblock {\em Distributed and Parallel Databases}, 18, 173--197.

\bibitem[Harding \& Banzhaf, 2007]{harding2007fast_ieee}
Harding, S. \& Banzhaf, W. (2007).
\newblock Fast genetic programming and artificial developmental systems on
  gpus.
\newblock {\em 21st International Symposium on High Performance Computing
  Systems and Applications (HPCS'07)}, 1--7.

\bibitem[Harris et~al., 2020]{harris2020array}
Harris, C.~R., Millman, K.~J., van~der Walt, S.~J., Gommers, R., Virtanen, P.,
  Cournapeau, D., Wieser, E., Taylor, J., Berg, S., Smith, N.~J., Kern, R.,
  Picus, M., Hoyer, S., van Kerkwijk, M.~H., Brett, M., Haldane, A., del
  R{\'{i}}o, J.~F., Wiebe, M., Peterson, P., G{\'{e}}rard-Marchant, P.,
  Sheppard, K., Reddy, T., Weckesser, W., Abbasi, H., Gohlke, C., \& Oliphant,
  T.~E. (2020).
\newblock Array programming with {NumPy}.
\newblock {\em Nature}, 585(7825), 357--362.
\newblock \url{https://doi.org/10.1038/s41586-020-2649-2}

\bibitem[Heinemann, 2008]{heinemann2008artificial}
Heinemann, C. (2008).
\newblock Artificial life environment.
\newblock {\em Informatik-Spektrum}, 31(1), 55--61.

\bibitem[Hernandez et~al., 2022]{hernandez2022can}
Hernandez, J.~G., Lalejini, A., \& Dolson, E. (2022).
\newblock What can phylogenetic metrics tell us about useful diversity in
  evolutionary algorithms?
\newblock {\em Genetic programming theory and practice XVIII}, 63--82.
  Springer.

\bibitem[Horgan, 1995]{horgan1995complexity}
Horgan, J. (1995).
\newblock From complexity to perplexity.
\newblock {\em Scientific American}, 272(6), 104--109.

\bibitem[Huerta-Cepas et~al., 2016]{huerta2016ete}
Huerta-Cepas, J., Serra, F., \& Bork, P. (2016).
\newblock Ete 3: reconstruction, analysis, and visualization of phylogenomic
  data.
\newblock {\em Molecular biology and evolution}, 33(6), 1635--1638.

\bibitem[Hunter, 2007]{hunter2007matplotlib}
Hunter, J.~D. (2007).
\newblock Matplotlib: A 2d graphics environment.
\newblock {\em Computing in Science \& Engineering}, 9(3), 90--95.
\newblock \url{https://doi.org/10.1109/mcse.2007.55}

\bibitem[Jia et~al., 2019]{jia2019dissecting}
Jia, Z., Tillman, B., Maggioni, M., \& Scarpazza, D.~P. (2019).
\newblock Dissecting the graphcore ipu architecture via microbenchmarking.
\newblock {\em arXiv preprint arXiv:1912.03413}.

\bibitem[Jouppi et~al., 2017]{jouppi2017datacenter}
Jouppi, N.~P., Young, C., Patil, N., Patterson, D., Agrawal, G., Bajwa, R.,
  Bates, S., Bhatia, S., Boden, N., Borchers, A., et~al. (2017).
\newblock In-datacenter performance analysis of a tensor processing unit.
\newblock {\em Proceedings of the 44th Annual International Symposium on
  Computer Architecture}, 1--12.

\bibitem[Kaplan et~al., 2020]{kaplan2020scaling}
Kaplan, J., McCandlish, S., Henighan, T., Brown, T.~B., Chess, B., Child, R.,
  Gray, S., Radford, A., Wu, J., \& Amodei, D. (2020).
\newblock Scaling laws for neural language models.
\newblock {\em arXiv preprint arXiv:2001.08361}.

\bibitem[Kim et~al., 2006]{kim2006discovery}
Kim, T.~K., Hewavitharana, A.~K., Shaw, P.~N., \& Fuerst, J.~A. (2006).
\newblock Discovery of a new source of rifamycin antibiotics in marine sponge
  actinobacteria by phylogenetic prediction.
\newblock {\em Applied and environmental microbiology}, 72(3), 2118--2125.

\bibitem[Klasky et~al., 2021]{osti_1770192}
Klasky, S., Thayer, J., \& Najm, H. (2021).
\newblock Data reduction for science: Brochure from the advanced scientific
  computing research workshop.
\newblock \url{https://doi.org/10.2172/1770192}

\bibitem[Krizhevsky et~al., 2012]{krizhevsky2012imagenet}
Krizhevsky, A., Sutskever, I., \& Hinton, G.~E. (2012).
\newblock Imagenet classification with deep convolutional neural networks.
\newblock {\em Advances in neural information processing systems}, 25,
  1097--1105.

\bibitem[Lalejini et~al., 2019]{lalejini2019data}
Lalejini, A., Dolson, E., Bohm, C., Ferguson, A.~J., Parsons, D.~P., Rainford,
  P.~F., Richmond, P., \& Ofria, C. (2019).
\newblock Data standards for artificial life software.
\newblock {\em ALIFE 2019: The 2019 Conference on Artificial Life}, 507--514.

\bibitem[Lalejini et~al., 2024a]{lalejini2024phylogeny}
Lalejini, A., Moreno, M.~A., Hernandez, J.~G., \& Dolson, E. (2024a).
\newblock Phylogeny-informed fitness estimation for test-based parent
  selection.
\newblock {\em Genetic Programming Theory and Practice XX}, 241--261. Springer
  International Publishing.
\newblock \url{https://doi.org/10.1007/978-981-99-8413-8_13}

\bibitem[Lalejini et~al., 2024b]{lalejini2024runtime}
Lalejini, A., Sanson, M., Garbus, J., Moreno, M.~A., \& Dolson, E. (2024b).
\newblock {\em Runtime phylogenetic analysis enables extreme subsampling for
  test-based problems}.
\newblock \url{https://doi.org/h10.48550/arXiv.2402.01610}

\bibitem[Langton, 1997]{langton1997artificial}
Langton, C.~G. (1997).
\newblock {\em Artificial life: An overview}.
\newblock Mit Press.

\bibitem[Lauterbach, 2021]{lauterbach2021path}
Lauterbach, G. (2021).
\newblock The path to successful wafer-scale integration: The cerebras story.
\newblock {\em IEEE Micro}, 41(6), 52--57.
\newblock \url{https://doi.org/10.1109/mm.2021.3112025}

\bibitem[Lemmon \& Lemmon, 2013]{lemmon2013high}
Lemmon, E.~M. \& Lemmon, A.~R. (2013).
\newblock High-throughput genomic data in systematics and phylogenetics.
\newblock {\em Annual Review of Ecology, Evolution, and Systematics}, 44,
  99--121.

\bibitem[Lenski et~al., 2003]{lenski2003evolutionary}
Lenski, R.~E., Ofria, C., Pennock, R.~T., \& Adami, C. (2003).
\newblock The evolutionary origin of complex features.
\newblock {\em Nature}, 423(6936), 139--144.

\bibitem[Lewinsohn et~al.,
  2023]{lewinsohnStatedependentEvolutionaryModels2023a}
Lewinsohn, M.~A., Bedford, T., Müller, N.~F., \& Feder, A.~F. (2023).
\newblock State-dependent evolutionary models reveal modes of solid tumour
  growth.
\newblock {\em Nature Ecology \& Evolution}, 7(4), 581--596.
\newblock \url{https://doi.org/10.1038/s41559-023-02000-4}.
\newblock Publisher: Nature Publishing Group

\bibitem[Liben-Nowell \& Kleinberg, 2008]{liben2008tracing}
Liben-Nowell, D. \& Kleinberg, J. (2008).
\newblock Tracing information flow on a global scale using internet
  chain-letter data.
\newblock {\em Proceedings of the national academy of sciences}, 105(12),
  4633--4638.

\bibitem[Lie, 2022]{lie2022cerebras}
Lie, S. (2022).
\newblock Cerebras architecture deep dive: First look inside the hw/sw
  co-design for deep learning: Cerebras systems.
\newblock {\em 2022 IEEE Hot Chips 34 Symposium (HCS)}, 1--34.

\bibitem[Loyola-Gonzalez, 2019]{loyola2019black}
Loyola-Gonzalez, O. (2019).
\newblock Black-box vs. white-box: Understanding their advantages and
  weaknesses from a practical point of view.
\newblock {\em IEEE access}, 7, 154096--154113.

\bibitem[Mahendran \& Vedaldi, 2015]{mahendran2015understanding}
Mahendran, A. \& Vedaldi, A. (2015).
\newblock Understanding deep image representations by inverting them.
\newblock {\em Proceedings of the IEEE conference on computer vision and
  pattern recognition}, 5188--5196.

\bibitem[Marcus, 2018]{marcus2018deep}
Marcus, G. (2018).
\newblock Deep learning: {A} critical appraisal.
\newblock {\em CoRR}, abs/1801.00631.
\newblock \url{http://arxiv.org/abs/1801.00631}

\bibitem[Medina \& Dagan, 2020]{medina2020habana}
Medina, E. \& Dagan, E. (2020).
\newblock Habana labs purpose-built ai inference and training processor
  architectures: Scaling ai training systems using standard ethernet with gaudi
  processor.
\newblock {\em IEEE Micro}, 40(2), 17--24.

\bibitem[Miikkulainen et~al., 2024]{miikkulainen2024evolving}
Miikkulainen, R., Liang, J., Meyerson, E., Rawal, A., Fink, D., Francon, O.,
  Raju, B., Shahrzad, H., Navruzyan, A., Duffy, N., et~al. (2024).
\newblock Evolving deep neural networks.
\newblock {\em Artificial intelligence in the age of neural networks and brain
  computing}, 269--287. Elsevier.

\bibitem[Moore, 2024]{moore2024cerebras}
Moore, S.~K. (2024).
\newblock {\em {C}erebras {U}nveils {I}ts {N}ext {W}aferscale {A}{I} {C}hip ---
  spectrum.ieee.org}.
\newblock \url{https://spectrum.ieee.org/cerebras-chip-cs3}.
\newblock [Accessed 08-04-2024].

\bibitem[Moreno, 2024a]{moreno2024hsurf}
Moreno, M.~A. (2024a).
\newblock {\em mmore500/hstrat-surface-concept}.
\newblock \url{https://doi.org/10.5281/zenodo.10779240}

\bibitem[Moreno, 2024b]{moreno2024pecking}
Moreno, M.~A. (2024b).
\newblock {\em mmore500/pecking}.
\newblock \url{https://doi.org/10.5281/zenodo.10701185}

\bibitem[Moreno et~al., 2022a]{moreno2022hereditary}
Moreno, M.~A., Dolson, E., \& Ofria, C. (2022a).
\newblock Hereditary stratigraphy: Genome annotations to enable phylogenetic
  inference over distributed populations.
\newblock Number~80 in The 2022 Conference on Artificial Life, 64.
\newblock \url{https://doi.org/10.1162/isal_a_00550}

\bibitem[Moreno et~al., 2022b]{moreno2022hstrat}
Moreno, M.~A., Dolson, E., \& Ofria, C. (2022b).
\newblock hstrat: a python package for phylogenetic inference on distributed
  digital evolution populations.
\newblock {\em Journal of Open Source Software}, 7(80), 4866.
\newblock \url{https://doi.org/10.21105/joss.04866}

\bibitem[Moreno et~al., 2023]{moreno2023toward}
Moreno, M.~A., Dolson, E., \& Rodriguez~Papa, S. (2023).
\newblock Toward phylogenetic inference of evolutionary dynamics at scale.
\newblock {\em ALIFE 2023: The 2023 Conference on Artificial Life}, volume
  Proceedings of the 2023 Artificial Life Conference of {\em Proceedings of the
  2023 Artificial Life Conference}, 79.
\newblock \url{https://doi.org/10.1162/isal_a_00694}

\bibitem[Moreno \& Ofria, 2022]{moreno2022exploring}
Moreno, M.~A. \& Ofria, C. (2022).
\newblock Exploring evolved multicellular life histories in a open-ended
  digital evolution system.
\newblock {\em Frontiers in Ecology and Evolution}, 10.
\newblock \url{https://doi.org/10.3389/fevo.2022.750837}

\bibitem[Moreno et~al., 2024a]{moreno2024guide}
Moreno, M.~A., Ranjan, A., Dolson, E., \& Zaman, L. (2024a).
\newblock {\em A guide to tracking phylogenies in parallel and distributed
  agent-based evolution models}.
\newblock \url{https://doi.org/10.48550/arXiv.2405.10183}

\bibitem[Moreno \& {Rodriguez Papa}, 2024]{moreno2024apc}
Moreno, M.~A. \& {Rodriguez Papa}, S. (2024).
\newblock {\em mmore500/alifedata-phyloinformatics-convert}.
\newblock \url{https://doi.org/10.5281/zenodo.10701178}

\bibitem[Moreno et~al., 2024b]{moreno2024algorithms}
Moreno, M.~A., {Rodriguez Papa}, S., \& Dolson, E. (2024b).
\newblock {\em Algorithms for efficient, compact online data stream curation}.
\newblock \url{https://arxiv.org/abs/2403.00266}

\bibitem[Moreno et~al., 2024c]{moreno2024analysis}
Moreno, M.~A., {Rodriguez Papa}, S., \& Dolson, E. (2024c).
\newblock {\em Analysis of phylogeny tracking algorithms for serial and
  multiprocess applications}.
\newblock \url{https://doi.org/10.48550/arXiv.2403.00246}

\bibitem[Moreno et~al., 2021a]{moreno2021case}
Moreno, M.~A., {Rodriguez Papa}, S., \& Ofria, C. (2021a).
\newblock Case study of novelty, complexity, and adaptation in a multicellular
  system.
\newblock {\em OEE4: The Fourth Workshop on Open-Ended Evolution}.
\newblock
  \url{http://workshops.alife.org/oee4/papers/moreno-oee4-camera-ready.pdf}

\bibitem[Moreno et~al., 2021b]{moreno2021conduit}
Moreno, M.~A., {Rodriguez Papa}, S., \& Ofria, C. (2021b).
\newblock Conduit: A c++ library for best-effort high performance computing.
\newblock {\em Proceedings of the Genetic and Evolutionary Computation
  Conference Companion}, Gecco '21, 1795--1800.
\newblock \url{https://doi.org/10.1145/3449726.3463205}

\bibitem[Moreno \& Yang, 2024]{moreno2024wse}
Moreno, M.~A. \& Yang, C. (2024).
\newblock {\em mmore500/wse-sketches}.
\newblock \url{https://doi.org/10.5281/zenodo.10779280}

\bibitem[Moreno et~al., 2024d]{moreno2024supplement}
Moreno, M.~A., Yang, C., Dolson, E., \& Zaman, L. (2024d).
\newblock {\em Supplemental materials for trackable agent-based evolution
  models at wafer scale}.
\newblock via Open Science Framework at \url{https://osf.io/bfm2z}.
\newblock \url{https://doi.org/10.17605/osf.io/bfm2z}

\bibitem[Murphy \& Ryan, 2008]{murphy2008simple}
Murphy, G. \& Ryan, C. (2008).
\newblock A simple powerful constraint for genetic programming.
\newblock {\em Genetic Programming}, 146--157.
\newblock \url{https://doi.org/10.1007/978-3-540-78671-9_13}

\bibitem[Neyman, 1971]{neyman1971molecular}
Neyman, J. (1971).
\newblock Molecular studies of evolution: a source of novel statistical
  problems.
\newblock {\em Statistical decision theory and related topics}, 1--27.
  Elsevier.

\bibitem[Ofria et~al., 2009]{ofria2009artificial}
Ofria, C., Bryson, D.~M., \& Wilke, C.~O. (2009).
\newblock Avida.
\newblock {\em Artificial Life Models in Software}, 3--35. Springer London.
\newblock \url{https://doi.org/10.1007/978-1-84882-285-6_1}

\bibitem[pandas developers, 2020]{reback2020pandas}
pandas developers (2020).
\newblock pandas-dev/pandas: Pandas.
\newblock {\em Zenodo}.
\newblock \url{https://doi.org/10.5281/zenodo.3509134}

\bibitem[Pattee, 1989]{pattee1989simulations}
Pattee, H.~H. (1989).
\newblock Simulations, realizations, and theories of life.
\newblock {\em IEEE Symposium on Artificial Life}.
\newblock \url{https://doi.org/10.4324/9780429032769}

\bibitem[Perumalla et~al., 2022]{perumalla2022computer}
Perumalla, K., Bremer, M., Brown, K., Chan, C., Eidenbenz, S., Hemmert, K.~S.,
  Hoisie, A., Newton, B., Nutaro, J., Oppelstrup, T., et~al. (2022).
\newblock Computer science research needs for parallel discrete event
  simulation (pdes).
\newblock Technical report, Lawrence Livermore National Lab.(LLNL), Livermore,
  CA (United States).

\bibitem[Ray, 1992]{ray1992evolution}
Ray, T. (1992).
\newblock Evolution, ecology and optimization of digital organisms.
\newblock {\em Santa Fe Institute working paper}, 92.
\newblock
  \url{https://homeostasis.scs.carleton.ca/~soma/adapsec/readings/tierra-92-08-042.pdf}

\bibitem[Ray, 1995]{ray1995proposal}
Ray, T. (1995).
\newblock A proposal to create a network-wide biodiversity reserve for digital
  organisms.
\newblock Technical Report Tr-h-133, Atr.
\newblock \url{http://tomray.me/pubs/reserves/}

\bibitem[Richmond et~al., 2010]{richmond2010high}
Richmond, P., Walker, D., Coakley, S., \& Romano, D. (2010).
\newblock High performance cellular level agent-based simulation with flame for
  the gpu.
\newblock {\em Briefings in bioinformatics}, 11(3), 334--347.

\bibitem[Schreiber et~al., 2021]{schreiber2021cross}
Schreiber, S.~J., Ke, R., Loverdo, C., Park, M., Ahsan, P., \& Lloyd-Smith,
  J.~O. (2021).
\newblock Cross-scale dynamics and the evolutionary emergence of infectious
  diseases.
\newblock {\em Virus evolution}, 7(1), veaa105.

\bibitem[Selig, 2022]{selig2022cerebras}
Selig, J. (2022).
\newblock The cerebras software development kit: A technical overview.

\bibitem[Shahbandegan et~al., 2022]{shahbandegan2022untangling}
Shahbandegan, S., Hernandez, J.~G., Lalejini, A., \& Dolson, E. (2022).
\newblock Untangling phylogenetic diversity's role in evolutionary computation
  using a suite of diagnostic fitness landscapes.
\newblock {\em Proceedings of the Genetic and Evolutionary Computation
  Conference Companion}, 2322--2325.

\bibitem[Stamatakis, 2005]{STAMATAKIS2005phylogenetics}
Stamatakis, A. (2005).
\newblock Phylogenetics: Applications, software and challenges.
\newblock {\em Cancer Genomics \& Proteomics}, 2(5), 301--305.
\newblock \url{https://cgp.iiarjournals.org/content/2/5/301}

\bibitem[Sutter et~al., 2005]{sutter2005free}
Sutter, H. et~al. (2005).
\newblock The free lunch is over: A fundamental turn toward concurrency in
  software.
\newblock {\em Dr. Dobb's journal}, 30(3), 202--210.

\bibitem[Taylor, 2019]{taylor2019evolutionary}
Taylor, T. (2019).
\newblock Evolutionary innovations and where to find them: Routes to open-ended
  evolution in natural and artificial systems.
\newblock {\em Artificial life}, 25(2), 207--224.

\bibitem[Taylor et~al., 2016]{taylor2016open}
Taylor, T., Bedau, M., Channon, A., Ackley, D., Banzhaf, W., Beslon, G.,
  Dolson, E., Froese, T., Hickinbotham, S., Ikegami, T., et~al. (2016).
\newblock Open-ended evolution: Perspectives from the oee workshop in york.
\newblock {\em Artificial life}, 22(3), 408--423.

\bibitem[Velten et~al., 2022]{markus2022memory}
Velten, M., Sch\"{o}ne, R., Ilsche, T., \& Hackenberg, D. (2022).
\newblock Memory performance of amd epyc rome and intel cascade lake sp server
  processors.
\newblock {\em Proceedings of the 2022 ACM/SPEC on International Conference on
  Performance Engineering}, Icpe '22, 165–175.
\newblock \url{https://doi.org/10.1145/3489525.3511689}

\bibitem[Virtanen et~al., 2020]{2020SciPy-NMeth}
Virtanen, P., Gommers, R., Oliphant, T.~E., Haberland, M., Reddy, T.,
  Cournapeau, D., Burovski, E., Peterson, P., Weckesser, W., Bright, J., {van
  der Walt}, S.~J., Brett, M., Wilson, J., Millman, K.~J., Mayorov, N., Nelson,
  A. R.~J., Jones, E., Kern, R., Larson, E., Carey, C.~J., Polat, {\.I}., Feng,
  Y., Moore, E.~W., {VanderPlas}, J., Laxalde, D., Perktold, J., Cimrman, R.,
  Henriksen, I., Quintero, E.~A., Harris, C.~R., Archibald, A.~M., Ribeiro,
  A.~H., Pedregosa, F., {van Mulbregt}, P., \& {SciPy 1.0 Contributors} (2020).
\newblock {{SciPy} 1.0: Fundamental Algorithms for Scientific Computing in
  Python}.
\newblock {\em Nature Methods}, 17, 261--272.
\newblock \url{https://doi.org/10.1038/s41592-019-0686-2}

\bibitem[Voznica et~al., 2022]{voznica2022deep}
Voznica, J., Zhukova, A., Boskova, V., Saulnier, E., Lemoine, F.,
  Moslonka-Lefebvre, M., \& Gascuel, O. (2022).
\newblock Deep learning from phylogenies to uncover the epidemiological
  dynamics of outbreaks.
\newblock {\em Nature Communications}, 13(1), 3896.

\bibitem[Waskom, 2021]{waskom2021seaborn}
Waskom, M.~L. (2021).
\newblock seaborn: statistical data visualization.
\newblock {\em Journal of Open Source Software}, 6(60), 3021.
\newblock \url{https://doi.org/10.21105/joss.03021}

\bibitem[{W}es {M}c{K}inney, 2010]{mckinney-proc-scipy-2010}
{W}es {M}c{K}inney (2010).
\newblock {D}ata {S}tructures for {S}tatistical {C}omputing in {P}ython.
\newblock {\em {P}roceedings of the 9th {P}ython in {S}cience {C}onference},
  56--61.
\newblock \url{https://doi.org/10.25080/Majora-92bf1922-00a}

\bibitem[Zhang et~al., 2016]{zhang2016cambricon}
Zhang, S., Du, Z., Zhang, L., Lan, H., Liu, S., Li, L., Guo, Q., Chen, T., \&
  Chen, Y. (2016).
\newblock Cambricon-x: An accelerator for sparse neural networks.
\newblock {\em 2016 49th Annual IEEE/ACM International Symposium on
  Microarchitecture (MICRO)}, 1--12.

\bibitem[Zhao \& Zhang, 2006]{zhao2005generalized}
Zhao, Y. \& Zhang, S. (2006).
\newblock Generalized dimension-reduction framework for recent-biased time
  series analysis.
\newblock {\em IEEE Transactions on Knowledge and Data Engineering}, 18(2),
  231--244.
\newblock \url{https://doi.org/10.1109/tkde.2006.30}

\end{thebibliography}


\begin{thebibliography}{}

\end{thebibliography}
